\documentclass[runningheads]{llncs}
\usepackage{graphicx}
\usepackage{comment}
\usepackage{amsmath,amssymb} % define this before the line numbering.
\usepackage{color}
\usepackage{epsfig}
\usepackage{amsmath}
\usepackage{amssymb}
\usepackage{balance}

\usepackage{enumitem}
\setlist{nosep} % or \setlist{noitemsep} to leave space around whole list
\setlist{itemsep=1pt, topsep=3pt}

\newcommand{\MS}[1]{\textcolor{blue}{\iffalse#1\fi}}
\newcommand{\SJ}[1]{\textcolor{black}{#1}}
\newcommand{\SSJ}[1]{\textcolor{red}{\iffalse#1\fi}}

\usepackage{multirow}
\usepackage{bm}
\usepackage{pifont}
\usepackage{footmisc} %%%\footref for refer a footnote.

\newcommand{\xmark}{\ding{53}} % x

\newcommand{\sset}[1]{\bm{\mathcal{\MakeUppercase{#1}}}} % Set
\newcommand{\smat}[1]{\bm{\MakeUppercase{#1}}}			% Matrix
\newcommand{\svec}[1]{\bm{\MakeLowercase{#1}}}			% Vector
\newcommand{\sfun}[1]{\mathcal{\MakeUppercase{#1}}}		% Function
\newcommand{\blfootnote}[2]{%
  \begingroup
  \footnote{\label{#1} #2}%
  \endgroup
}
\usepackage[pagebackref=true,breaklinks=true,letterpaper=true,colorlinks,bookmarks=false]{hyperref}

\begin{document}
% \renewcommand\thelinenumber{\color[rgb]{0.2,0.5,0.8}\normalfont\sffamily\scriptsize\arabic{linenumber}\color[rgb]{0,0,0}}
% \renewcommand\makeLineNumber {\hss\thelinenumber\ \hspace{6mm} \rlap{\hskip\textwidth\ \hspace{6.5mm}\thelinenumber}}
% \linenumbers
\pagestyle{headings}
\mainmatter
\def\ECCVSubNumber{4729}  % Insert your submission number here

%\title{Simultaneous Detection and Tracking of Multiple objects in Video } % Replace with your title
\title{Simultaneous Detection and Tracking with Motion Modelling for Multiple Object Tracking }

% INITIAL SUBMISSION 
%\begin{comment}
% \titlerunning{ECCV-20 submission ID \ECCVSubNumber} 
% \authorrunning{ECCV-20 submission ID \ECCVSubNumber} 
% \author{Anonymous ECCV submission}
% \institute{Paper ID \ECCVSubNumber}
%\end{comment}
%******************

% CAMERA READY SUBMISSION
% \begin{comment}
\titlerunning{Deep Motion Modeling Network for MOT}
% If the paper title is too long for the running head, you can set
% an abbreviated paper title here
%
\author{ShiJie Sun\inst{1} \and
Naveed Akhtar\inst{2} \and
XiangYu Song\inst{3} \and
HuanSheng Song\inst{1} \and
Ajmal Mian\inst{2} \and
Mubarak Shah\inst{4}
}
\authorrunning{S. Sun, N. Akhtar, et al.}
% First names are abbreviated in the running head.
% If there are more than two authors, 'et al.' is used.
%
\institute{
Chang'an University, Xi'an, Shaanxi, China \\
\email{\{shijieSun,hshsong\}@chd.edu.cn} \and
University of Western Australia, 35 Stirling Highway, Crawley, WA, Australia \\
\email{\{naveed.akhtar,ajmal.mian\}@uwa.edu.au} \and
%\url{http://www.springer.com/gp/computer-science/lncs} \and
Deakin University, RWaurn Ponds, Victoria 3216, Melbourne, Australia \\
\email{xiangyu.song@deakin.edu.au}\and
University of Central Florida, Orlando, FL, America \\
\email{shah@crcv.ucf.edu}
}
% \end{comment}
%******************
\maketitle

\begin{abstract}
  %\MS{I am not sure "Deep Motion Modeling" in the title  is necessary here, in the paper that point is not that obvious, remove those words?}\\
  %\SJ{I've removed the ``Deep Motion Modeling'' in the title.}
  Deep learning based Multiple Object Tracking (MOT) currently relies on off-the-shelf detectors for tracking-by-detection. This results in deep models that are detector biased and evaluations that are detector influenced. To resolve this issue, we introduce Deep Motion Modeling Network (DMM-Net) that can estimate multiple objects' motion parameters to perform joint detection and association in an end-to-end manner. DMM-Net models object features over multiple frames and simultaneously infers object \SJ{classes}, visibility and their motion parameters. These outputs are readily used to update the tracklets for efficient MOT. DMM-Net achieves PR-MOTA score of 12.80 @ 120+ fps for the popular UA-DETRAC challenge - which is better performance and orders of magnitude faster. We also contribute a synthetic large-scale public dataset Omni-MOT for vehicle tracking that provides precise ground-truth annotations to eliminate the detector influence in MOT evaluation. This 14M+ frames dataset is extendable with our public script (Code at \href{https://github.com/shijieS/OmniMOTDataset}{Dataset}, \href{https://github.com/shijieS/OMOTDRecorder}{Dataset Recorder}, \href{https://github.com/shijieS/DMMN}{Omni-MOT Source}). We demonstrate the suitability of Omni-MOT for deep learning with DMM-Net, and also make the source code of our network public.
\keywords{Multiple Object Tracking, Tracking-by-detection, Deep Learning, Simultaneous detection and tracking.}
\end{abstract}

%\vspace{-7mm}
\section{Introduction}
 \label{sec:Intro}
   %\vspace{-3mm}
 Multiple Object Tracking (MOT)~\cite{dehghan2015gmmcp}, \cite{Leal-Taixe2015}, \cite{Sun2018b},  \cite{voigtlaender2019mots} is a longstanding problem in Computer Vision~\cite{Luo2017}.
% It aims at simultaneously  detecting multiple objects in individual video frames and associating them over extended periods of time. I am removing this sentence, it is not correct and confusing.
 Contemporary deep learning based MOT has widely adopted the tracking-by-detection paradigm~\cite{Tian2018}, that capitalizes on the natural division of {\em detection} and {\em data association} tasks for the problem. In standard MOT evaluation protocol, the object detections are assumed to be known and public detections are provided on evaluation sequences, and MOT algorithms are expected to output object tracks by solving the data association problem. 
 %This adoption is largely influenced by the progress in object detection.  I am removing this sentence
 %Each with its own contrivance, in particular, deep learning based detectors~\cite{Ren2017a,Liu2016,Redmon2018,dai2016r} are becoming increasingly better at  this task.
 
 Although attractive, using off-the-shelf detectors for MOT also has undesired ramifications. For instance, a deep model employed for the subsequent data association task (or a constituent sub-task) gets biased to the detector. The detector performance can also become a bottle-neck for the overall tracker. Additionally, composite techniques resulting from independent detectors fail to harness the true representation power of deep learning by sacrificing end-to-end training etc. 
%Besides these issues, current tracking-by-detection paradigm also struggles with fair evaluation of tracking techniques. Detector biased ground-truth for large-scale datasets makes it hard to ascertain transparent evaluation. Not to mention, the complex task of MOT already requires a long list of metrics for correct performance interpretation.  All these problems are potentially solvable if trackers can implicitly detect the target objects, and detector bias can be removed from the ground-truth labeling of the tracks. This work makes a stride towards these solutions.
 %%\SJ{Besides these issues, the current evaluation of multiple object tracking techniques heavily depend on off-the-shelf detectors. Detector biased evaluation does promote the MOT community and does not pave the way to a fair challenge. However, to some extent, it also makes some researchers focus too much on fixing detector problems as compared to developing general tracking solutions. 
 %the combination complicated  (but not generalizable) tricks.
 It also seems unnatural that a MOT tracker must be evaluated on different detectors to interpret its tracking performance.
 All these problems are potentially solvable if trackers can implicitly detect the target objects, and detector bias can be removed from the ground-truth labeling of the tracks. This work makes strides towards these solutions.
 
  \begin{figure*}[t]
   \centering
   \includegraphics[width=\linewidth]{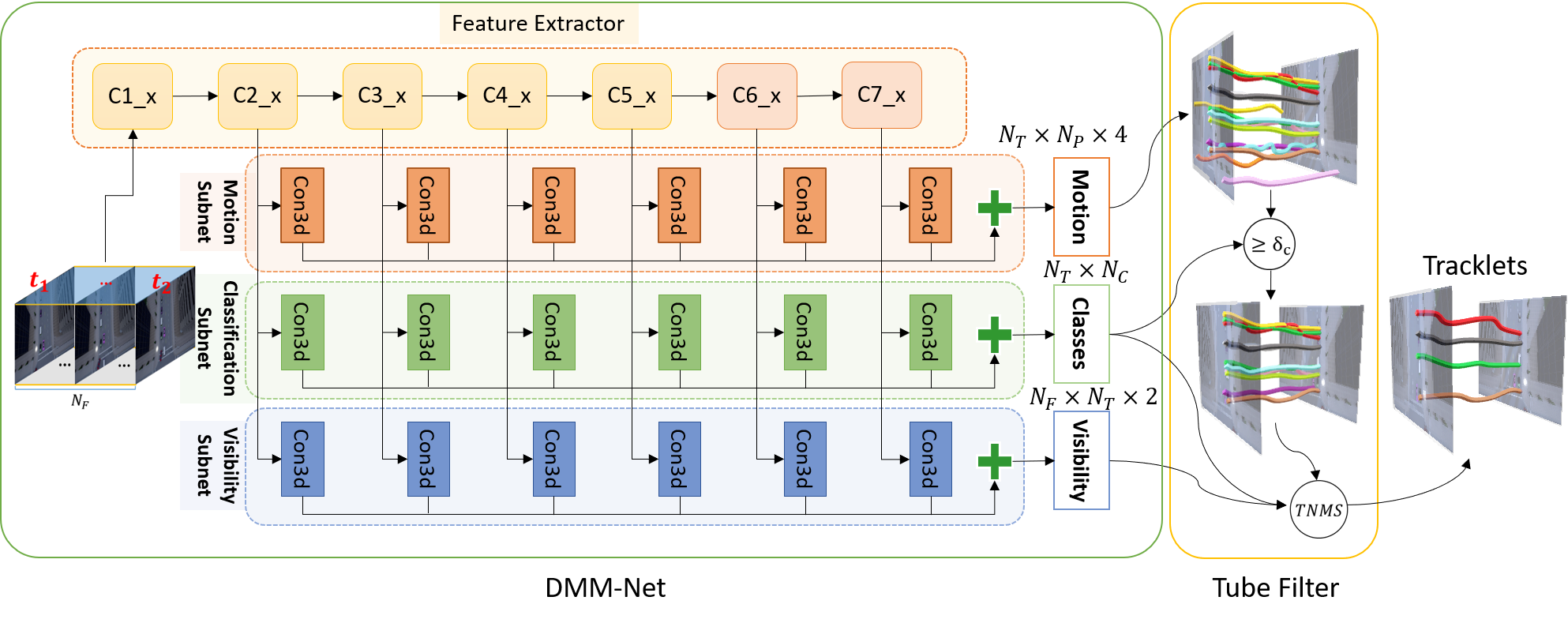}
   \caption{Schematics of the proposed end-to-end trainable DMM-Net: $N_F$ frames from time stamp ${t_1:t_2}$ are input to the network. The frame sequence is first processed with a \textit{Feature Extractor} comprising 3D ResNet-like~\cite{Hara2018} convolutional groups. Outputs of the last groups 2 to 7 are processed by Motion Subnet, Classifier Subnet\MS{it is better to use classification instead of categories used later}, and Visibility Subnet. Each sub-network uses 3D convolutions to learn features that are concatenated to predict anchor \SJ{tubes'} motion parameters ($\smat O_M \in \mathbb{R}^{N_T\times N_P \times 4}$), object \SJ{classes} \MS{classes}  ($\smat O_C \in \mathbb{R}^{N_T\times N_C}$), and visibility ($\smat O_V \in \mathbb{R}^{N_F \times N_T \times 2}$), where $N_T$, $N_P$ and $N_C$ denote the number of anchor \SJ{tubes}, motion parameters and object \SJ{classes (including background)}. We explain anchor tubes and motion parameters in Section~\ref{sec:fmmn}. DMM-Net is trained with its specialized loss. For deployment, {\color{black} the anchor tubes predicted by the DMM-Net are filtered to compute tracklets defined over multiple frames. These tracklets are later combined to form a complete track. }}

   \label{fig:pipeline}
   %\vspace{-3mm}
\end{figure*}

 We make two major contributions towards setting the tracking-by-detection paradigm free from off-the-shelf detectors in deep learning settings.
 Our first contribution comes as the first-of-its-kind deep network that performs object detections and data association \SJ{by estimating multiple object motion parameters} in an end-to-end manner. Our network, called Deep Motion Modeling Network (DMM-Net),  predicts object motion parameters, their \SJ{classes}  and their visibility with the help of three sub-networks, see Fig.~\ref{fig:pipeline}. These sub-networks exploit feature maps of frames in a video sequence that are learned with a Feature Extractor comprising seven 3D ResNet-like~\cite{Hara2018} convolutional groups.
 Instead of individual frames, DMM-Net simultaneously processes multiple (i.e.~16) frames.
 To handle multiple tracks in those frames, we introduce the notion of anchor \SJ{tubes} %\MS{I had hard time understanding tunnels first; now I think you mean tubes, which is widely used, why do not you used tubes?}
 that extends the concept of anchor boxes in object detection~\cite{Liu2016} along the temporal dimension for MOT. \SJ{Similar to~\cite{Liu2016}, these anchor tubes are pre-defined to reduce the computation and improve the network speed. The predicted motion parameters can describe the shape offset and scale of each pre-defined anchor tube in the spatio-temporal space.}
 We propose individual losses over the comprehensive representations of the sub-networks to predict object motion parameters\MS{what are object motion parameters, give an idea}, object \SJ{classes} and visibility. 
 The DMM-Net output is readily usable  to update the tracks. 
 As our second major contribution, we propose a realistic large-scale dataset with accurate and extensive ground-truth annotations.  The proposed dataset, termed Omni-MOT for its comprehensive coverage of the MOT scenarios, is generated with CARLA simulator~\cite{Koltun2017} for vehicle tracking. The dataset comprises 14M+ frames, 250K tracks, 110 million bounding boxes, three weather conditions, three crowd levels and three camera views in five simulated towns. By eliminating the use of off-the-shelf detectors in ground-truth labeling, it removes any detector bias in evaluating the techniques.
 
 The Omni-MOT dataset and  DMM-Net source code are both publicly available for the broader research community. For the former, we also provide software tools to freely extend the dataset. We demonstrate the suitability of the Omni-MOT for deep models by training and evaluating DMM-Net on its videos. We also augment DMM-Net with Omni-MOT and evaluate our technique on the popular UA-DETRAC challenge~\cite{Wen2015a}. The remote server computed results show that DMM-Net is able to achieve a very competitive 12.80 score for the comprehensive PR-MOTA metric with the overall speed of 123 fps.
 The orders of magnitude increase in the speed is directly attributed to the intrinsic detections in our tracking-by-detection technique.

 \section{Related Work}\label{sec:rw}
 %\vspace{-3mm}
 
%  Multiple object tracking is the primary task of computer vision and has achieved significant progress with the development of deep learning.
%  In this section, we first introduce some popular tracking approaches. After that, we review some multiple object tracking datasets.

%  \subsection{Multiple Object Trackers}

Multiple Object Tracking (MOT) is a fundamental problem in  Computer Vision that has attracted significant interest of researchers in recent years.
For a general review of the related literature, we refer to Luo et al.~\cite{Luo2017} and Emami et al.~\cite{Emami2018}. Here, we focus on the key contributions that relate to this work more closely.
With the recent advances  in object detectors, tracking-by-detection is fast becoming the common contemporary paradigm for MOT~\cite{shafique2005noniterative}, \cite{sheng2018heterogeneous}, \cite{Tian2018}, \cite{wen2018learning}.
In this scheme, objects are first detected frame-wise and later associated with each other across multiple frames. 
Relying on off-the-shelf detectors, techniques following this paradigm mainly focus on object association, which can make them inherently detector biased.  %the association procedure can inherently get biased to the used detector. 
These methods can be broadly categorized as local~\cite{reid1979algorithm}, \cite{shu2012part} and global~\cite{dehghan2015gmmcp}, \cite{RoshanZamir2012}, \cite{wu2007detection}  approaches.
The former use two frames for data association, while the latter associate objects over a larger number of frames.

More recent global techniques cast data association into a network flow problem \cite{Berclaz2011}, \cite{butt2013multi}, \cite{Pirsiavash2011a},  \cite{shitrit2014multi}.
For instance, Berclaz et al.~\cite{Berclaz2011} solved a constrained flow optimization problem for MOT and used the k-shortest paths algorithm for associating the tracks.
Chari et al.~\cite{Chari2015} added a pairwise cost to the min-cost network flow framework and proposed a convex relaxation of the problem.
However, such methods rely on object detectors rather strongly, which makes them less attractive in the presence of occlusions and misdetections.
To mitigate the problems resulting from occlusions, 
%Shu et al.~\cite{shu2012part} extended a part based on human detector~\cite{felzenszwalb2010object}.
Milan et al.~\cite{Milan2014} employed a continuous energy minimization framework for MOT that incorporates explicit occlusion reasoning and appearance modeling.
Wen et al.~\cite{Wen2014} also proposed a data association technique based on undirected hierarchical relation hyper-graph.

%Recent years have witnessed numerous deep learning based  techniques for MOT that use pre-trained networks to build a feature model.
%Object features computed by the model are later used for data association. 
%Improving this scheme, Schulter et al.~\cite{Schulter2017} proposed a network that is trained for data association in the context of MOT.
%Son et al.~\cite{Son2017} proposed a quadruplet CNN that learns to associate objects detected in different video frames. 
Sun et al.~\cite{Sun2018b} proposed a deep affinity network to model features of pre-detected objects and compute object affinities by the same network.
Bea et al.~\cite{Baea} modified the Siameses Network to learn deep representations for MOT with object association. 
%They combined online transfer learning with a modified network to fine-tune the latter for online tracking.
There are few instances of deep learning techniques that aim at removing the reliance of tracking on independent detectors. For instance, Feichtenhofer et al~\cite{Feichtenhofer2017} proposed an R-FCN based network~\cite{dai2016r} that performs object detection and jointly builds an object model to be used for data association.
However, their method is limited to frame-wise detections. Consequently, it only allows frame-by-frame association, requiring manual adjustment of temporal stride. Our technique is inherently different, as it directly computes tracklets over multiple frames by motion modeling, enabling realtime solutions while considering all the frames. 
Besides the development of novel techniques, the role of datasets is central to the advancement of deep learning based MOT. 
%The dataset of is multiple object tracking crucial for the deep-learning-based techniques. 
Currently, a few large datasets for this task are available, e.g.~PETS2009~\cite{Ferryman2009}, KITTI~\cite{Geiger2012}, DukeMTMC~\cite{RistaniErgysandTomasi2014}, PoseTrack~\cite{Iqbal2017}, and MOT Challenge datasets~\cite{Leal-Taixe2015}, \cite{MilanL0RS16}.
These datasets are recorded in the real world with pedestrians and vehicles as the objects of interest. We refer to the original works for more details on these datasets. Below, we briefly discuss UA-DETRAC~\cite{Wen2015a} for its high relevance to our contribution. 
 
 % MOT 17 Dataset
%MOT Challenge Dataset (MOT15~\cite{Leal-Taixe2015}, MOT16~\cite{MilanL0RS16}, MOT17~\cite{MilanL0RS16}) is one of the most popular multiple objects tracking dataset. It plays a critical role in the development of MOT. It mainly focuses on tracking pedestrians and contains different types of cameras (static camera, moving camera ) and different viewpoints (low viewpoint, elevated viewpoint). However, the scale of this dataset is not that large. Besides, the bounding boxes of the training dataset are manually labeled while the bounding boxes of the testing dataset are labeled by the detector (i.e., DPM~\cite{Felzenszwalb2008a}, SDP~\cite{Yang2016} or FRCNN~\cite{Ren2017a}).
 
 % Introduce the CARLA
 
UA-DETRAC~\cite{Wen2015a} is a large dataset for traffic scenes MOT. It provides object bounding boxes, their IDs and information on the overlapped ratio of the objects. However, the provided detections are individually generated by detectors and hence are prone to errors. This results in an undesired detector-bias in tracker evaluation.  %In this condition, the detection-based tracker performs well on the training dataset and worse on the testing dataset. 
Besides, different pre-processing procedures of the detectors employed by the dataset also cause problems in fair evaluation. 
%Besides, different pre-processing methods of detection may cause the unfairness of the competition. 
Although UA-DETRAC has served a great purpose in advancing the state-of-the-art in MOT for vehicles, the aforementioned issues call for a more transparent dataset that does not rely on off-the-shelf detectors for evaluating trackers.
This work provides such a dataset with realistic settings and complete control over the environment conditions and camera viewpoints. 
%The proposed large-scale  dataset is also extendable and can further be used to augment the real-world datasets for improved tracking performance.  

%We publish a virtual yet large scale dataset containing a variety of vehicles and weather conditions. 
 %Different from other datasets, this dataset can be easily extended. 
 %We provide the primary dataset, which contains a training set and a testing set. 
 %The detections of these two sets are consistent. 
 %As far as we know, this dataset is the first large scale dataset for the MOT task generated by the 3D simulator.

 %------------------------------------------------------------------------
  %\vspace{-5mm}
 \section{Omni-MOT Dataset}\label{sec:d}
 %\vspace{-3mm}
 We term the proposed dataset as \textbf{O}mni-\textbf{MOT} (OMOT) dataset for its comprehensive coverage of the conditions and scenarios possible in MOT. The dataset is publicly available for download. Moreover, we also make the recording script for the dataset public that will enable the community to further extend the data
 %\footnote{All embedded URLs in the Abstact will be made public after acceptance.}. 
 The provided script has the ability to generate multi-camera videos. The dataset is recorded using virtual cameras in the CARLA simulator~\cite{Koltun2017}.

%%%%%%%%%%%%%%% MOVE TO SUPP%%%%%%%%%%%% 
 %In the proposed dataset, five different simulated cities are considered. For each city, we use up to 39 cameras. The cameras are placed with viewpoints that have three levels of difficulty for the MOT scenarios. Namely, (a) Easy view: which results in no occlusion of the vehicles. (b) Ordinary view: that allows temporary occlusions but forbids continuous occlusions. (c) Hard view: that allows continuous occlusions in the videos. Collectively, we provide $90$ scenes in the dataset that result in $3,510$ videos. There are $14.04$M frames of size $1920 \times 1080$ in the OMOT dataset that are recorded in the `XDIV' format to provide high-quality videos with acceptable memory size.   
 
 %Besides controlling the camera viewpoints, we include three different weather conditions; clear, cloudy and rainy. We also include three levels of traffic congestion in the OMOT dataset i.e. low, medium and severe congestion. We provide further details of the dataset in the supplementary material, where we  also include a video showing different scenes, camera viewpoints and weather conditions. CARLA simulator allows us to capture comprehensive information on the target objects with high precision. Hence, besides being accurate, our ground truth annotations are detailed enough to be used for other related problems, e.g.~velocity estimation, 3D localization, camera calibration. 
%%%%%%%%%%%%%%% MOVE TO SUPP%%%%%%%%%%%% 

To the best of our knowledge, Omni-MOT is the first realistic large dataset for tracking vehicles that completely relinquishes off-the-shelf detectors in ground truth generation, and provides comprehensive annotations. Moreover, with the provided scripts, the dataset can easily be extended for future research.
To put the scale of Omni-MOT into perspective, the provided number of frames  is almost $1,200$ times larger than MOT17. The number of provided tracks and boxes respectively are $210$ and $30$ times larger than UA-DETRAC. Not to mention, all the boxes and tracks are simulator generated that avoids any  labeling error. Considering that OMOT can also be used for data augmentation, we  make the ground truth for the test videos public as well. Please see the supplementary material of the paper for complete details of the dataset.

 \section{\textbf{D}eep \textbf{M}otion \textbf{M}odeling Network}\label{sec:fmmn}
 %\vspace{-2mm}
 To absolve deep learning based tracking-by-detection from independently pre-trained off-the-shelf detectors, we propose \textbf{D}eep \textbf{M}otion \textbf{M}odeling \textbf{Net}work (DMM-Net) for online MOT (see Fig.~\ref{fig:pipeline}). 
 Our network enables MOT by jointly performing object detection, tracking, and \SJ{classification} across multiple video frames without requiring pre-detections and subsequent data association. For a given input video, it outputs objects' motion parameters, \SJ{classes}, and their visibility across the input frames.
 We provide a detailed discussion of our network below. 
However, we first introduce the used notations and conventions. % for a concise description of the text and figures.
 %We strive to explore the power of deep learning to perform online end-to-end tracking. Central to our technique is an CNN-based \textbf{F}ast \textbf{M}otion \textbf{M}odel \textbf{N}etwork (FMMN) as shown in Fig.~\ref{fig:pipeline},
 %which can learn the spatial-temporal features for jointly performing detection, tracking, and categorizing across a set of video frames without doing association.
 %It outputs all possible objects' motion parameters, \SJ{classes}, and visibilities over the input video frames. 
 %We provide a detailed discussion of our network below. 
 %Before that, we firstly introduce notations and conventions for a concise description of the text and figures.
 
\begin{figure}[t]
 \centering
 \includegraphics[width=0.8\textwidth]{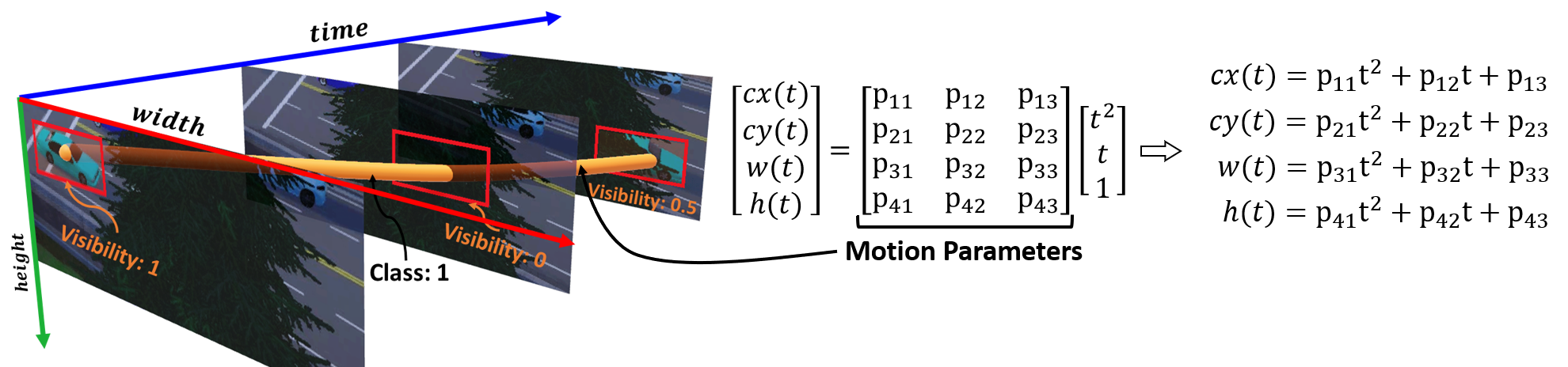}
 \caption{(Best viewed in color) Illustration of motion visibility, class, and motion parameters: The visibility of the vehicle goes from 1 (fully visible) to 0 (invisible) and back to 0.5 (partially visible). Classes are predefined as 1 for vehicle and 0 for everything else. Motion parameters \{$p_{11}, \cdots, p_{43}$\} are used to locate the center x ($cx$), center y ($cy$), width ($w$) and height ($h$) of the tube/tracklet at anytime. We employ a quadratic motion model leveraging $4\times 3$ matrices.}
 %\vspace{-1mm}
 \label{fig:motion_parameters}
 \end{figure}

 %\vspace{1mm}
 %\noindent\textbf{Notations:}
 %\vspace{-2.7mm}
 \begin{itemize}
     \item $N_F, N_C, N_P, N_T$  denote the number of input frames, object \SJ{classes} (0 for `background'), time-related motion parameters, and anchor \SJ{tubes}.
    \item $W, H$ are the frame width, and frame height.
    \item $\smat I_t$ denotes the video frame at time $t$. Subsequently, a 4-D tensor $\smat I_{t_1:t_2:N_F}\in \mathbb{R}^{3\times N_F \times W \times H}$ denotes  $N_F$ video frames from time $t_1$ to $t_2-1$. For simplicity, we often ignore the subscript ``$:N_F$''.
    \item $\smat B_{t_1:t_2:N_F},  \smat C_{t_1:t_2:N_F}, \smat V_{t_1:t_2:N_F}$ respectively denote the ground truth boxes, \SJ{classes} %\MS{do you mean classes?},
    and visibilities in the selected $N_F$ video frames from time $t_1$ to $t_2-1$. The text also ignores ``$:N_F$'' for these notations.
    \item $\smat O_{M, t_1:t_2:N_F}, \smat O_{C, t_1:t_2:N_F}, \smat O_{V, t_1:t_2:N_F}$ denote the estimated motion parameters, \SJ{classes}, and visiblities. With time stamps and frames clear from the context, we simplify these notations as $O_M, O_C, O_V$.
    In Fig.~\ref{fig:motion_parameters}, we illustrate object visibility, classes and motion parameters. %{\color{red}Fig.~\ref{fig:motion_parameters} can provide an intuition of visibility, class and motion parameters.}
   %  {\color{green} }{\color  {blue} All other notations except visibility, classes and motion parameters are standard. It will be good if, here you give more details about motion parameters i.e. you assume quadratic motion of width, height, x center, y center, and estimate three parameters. Similarly mention what are the classes, are they different cars, or just car class and background, and how many total classes. Also, mention what does visibility represents how it is determined? That will help the reviewers to understand the paper. Right now reading the paper quickly I do not have full idea about these.}%What are the motion parameters? Reviewers as I am curious what exactly you are estimating? it is speed, acceleration, or what?}
 \end{itemize}

 %\vspace{-1mm}
 \noindent\textbf{Conventions:}
 The shape of a network output tensor is considered to be
 Batch $\times$ Channels $\times$ Duration $\times$ Width $\times$ Height, where Duration accounts for the number of frames. For brevity, we often ignore the Batch dimension. %\MS{do you want to use notations $W$, $H$ and $N_F$ defined above?} \SSJ{($W, H, N_F$ represents the input frame shape. We use the ``Width'' and ``Height'' to indicate the format of output tensor size in the network.)}. 
 %$\mathop{Batch} \times \mathop{Channel} \times \mathop{Duration} \times \mathop{Width} \times \mathop{Height}$, where $\mathop{Duration}$ is the frame number along the temporal dimension. For purpose of brevity, we often leave out the $\mathop{Batch}$ dimension.
 
 % \begin{equation}
 %     \left\{\begin{matrix}
 %     u = \frac{p_1t+p_2}{p_3t + 1} \\
 % v = \frac{p_5t+p_6}{p_7t + 1}
 % \end{matrix}\right.
 % \end{equation}

 %\vspace{-3mm}
 \subsection{Data Preparation}
 \label{sec:DP}
 %\vspace{-1mm}
 Appropriate data preparation is important for training (and later testing) our network. In this process, we also avail the opportunity of augmenting the data to induce robustness in our model against camera photometric distortions, background scene variations, and other practical factors. 
 The inputs expected by our network are summarized in Table~\ref{tab:input_item}. For training a robust network, we perform the following data pre-processing:
 
  \begin{table}[t]
 \begin{center}
    \small
       \caption{Inputs for training and testing DMM-Net. The tensor dimensions are given as Channels$\times$Duration$\times$Width$\times$Height. $n_{t_1:t_2}$ is the number of tracks in $I_{t_1:t_2}$.}
       %\vspace{-2mm}
    \begin{tabular}{lccc}
    \hline
    Input               &        Dimensions/Size                    &   Train   &   Test\\ 
    \hline
    $\smat I_{t_1:t_2}$ &        $ 3 \times N_F \times W \times H$  &   \checkmark  &   \checkmark\\
    $\svec t_{t_1:t_2}$ &        $ N_F $                            &   \checkmark  &   \checkmark\\
    $\smat B_{t_1:t_2}$ &        $N_F \times n_{t_1:t_2} \times 4 $ &   \checkmark  &   \xmark\\
    $\smat C_{t_1:t_2}$ &        $n_{t_1:t_2} $ &   \checkmark  &   \xmark\\
    $\smat V_{t_1:t_2}$ &        $N_F \times n_{t_1:t_2} $ &   \checkmark  &   \xmark\\
    \hline
    \end{tabular}
    \label{tab:input_item}
 \end{center}
 %\vspace{-5mm}
 \end{table}

 %The provided ground truth format of most MOT datasets does not fit our network. Besides, the dataset should contain various camera photometric distortion, background scene variations, and other practical factors to train network more robust. Hence, we first introduce the input data format and then perform the preprocessing on these input data.
 
 %As given in Tab.~\ref{tab:input_item}, There are two input items: sequential video frames and 0-based time stamps. In the training phase, to calculate the loss of the network, there are three more input items: bounding boxes, each object's classification, and visibility. Note that, the input items in the training phase needs the following preprocessing:
 
 %\vspace{-2mm}
 \begin{enumerate}
    \item \textit{Stochastic frame sequence}  is introduced by randomly choosing $N_F$ frames from $2N_F$ %\SJ{consecutive} %\MS{do you mean consecutive, then use this word} 
    frames.
    \item \textit{Static scene emulation} is done by duplicating selected frames $N_F$  times.
    \item \textit{Photometric camera distortions} are introduced in the frames by scaling each pixel  by a random value in the range [0.7, 1.5]. The resulting frames are converted to  HSV format, and their  saturation channel is again scaled by a random value in [0.7, 1.5]. A frame is then converted back to RGB format and re-scaled by a random value in the same range. This process of photometric distortion is inspired  by~\cite{Sun2018b}.
    \item \textit{Frame expansion} is performed by enlarging video frames by a random factor in the range [1, 1.2]. To that end, we pad the original frames with extra pixels whose value is set to the mean pixel value of the training data.
    %\item \textit{Ground truth matrix generating}: The zero matrix $\smat B_{t_1:t_2},  \smat C_{t_1:t_2}, \smat V_{t_1:t_2}$ is filled by the detected boxes, \SJ{classes} and visibility in video frames from $t_1$ to $t_2-1$. 
 \end{enumerate}
 %\vspace{-1mm}   
 In our pre-processing, the above-mentioned steps 2-4 are applied to the frames with probability 0.01, 0.5, 0.5, respectively. The resulting frame are then resized to the fixed dimension $H\times W \times 3$, and horizontally flipped with a probability of 0.5. We simultaneously process the selected $N_F$ frames by applying the above mentioned transformations. These RGB video frames are then arranged as 4D-tensors $\smat I_{t_1:t_2}  \in \mathbb{R}^{3 \times N_F \times W \times H}$.  
We fill the ground truth data matrices $\smat B_{t_1:t_2}$,  $\smat C_{t_1:t_2}$, $\smat V_{t_1:t_2}$  with the detected boxes, %classes  \MS{at this point of the paper, I still do not know what are categories or classes? Different vehicles?, it will be good to explain this} 
  and visibilities in the video frames from $t_1$ to $t_2-1$.  We set the labeled boxes, classes, and visibilities of fully occluded boxes to 0. We also remove the tracks whose ratio of fully occluded boxes are greater than $\delta_v$. We let $N_F=16$ in our experiments. 
 \subsection{Anchor \SJ{Tubes}}
 \label{sec:AT}
 %\vspace{-3mm}
 Inspired by the effective Single Shot Detector (SSD)~\cite{Liu2016} for object detection, we extend the core concept of this technique for MOT. 
 Analogous to the anchor boxes of SSD, we introduce  \textit{anchor \SJ{tubes}} for DMM-Net. 
 Here, an anchor \SJ{tube} is a set of anchor boxes (and associated object \SJ{class} and visibility), that share the same location in multiple frames along the temporal dimension (see supplementary material for further  visualization).
 The information of anchor \SJ{tubes} is encoded with the tensors $\smat B_{t_1:t_2}$,  $\smat C_{t_1:t_2}$, and $\smat V_{t_1:t_2}$. The DMM-Net is designed to predict the \SJ{tube} shape offset parameters along the temporal dimension, the confidence for each object class, and the visibility of each box in the \SJ{tube}.

 %We propose the \textbf{anchor \SJ{tubes}} by extending the anchor boxes (or default boxes) from SSD~\cite{Liu2016}. 
 %An anchor tunnel is a set of anchor boxes along with category and visibility, which share the same location in the selected frames, as shown in Fig.~2 in the supplementary material.
 %They are described by three matrices $\smat B_A \in \mathbb{R}^{N_F \times N_T \times 4}, \smat C_A \in \mathbb{R}^{N_T}, \smat V_A \in \mathbb{R}^{N_F \times N_T}$.
 %The anchor \SJ{tubes} can be employed by the network to predict all possible objects' motion parameters, categories, and visibilities.
 %Our network predicts the shape offset parameters relative to time, the confidence for each category, and the visibility of each box in the anchor tunnel. 
%  Fig.~\ref{fig:anchor_tunnels} shows the difference between anchor boxes and anchor \SJ{tubes}. Similar to anchor boxes, the anchor \SJ{tubes} target to formatting the output of the network. 
 %For each anchor tunnel, our network predicts the shape offset parameters relative to the time, the confidence for each category, and the visibility of each box. 

\begin{figure}[t]
 \centering
 \includegraphics[width=0.8\textwidth]{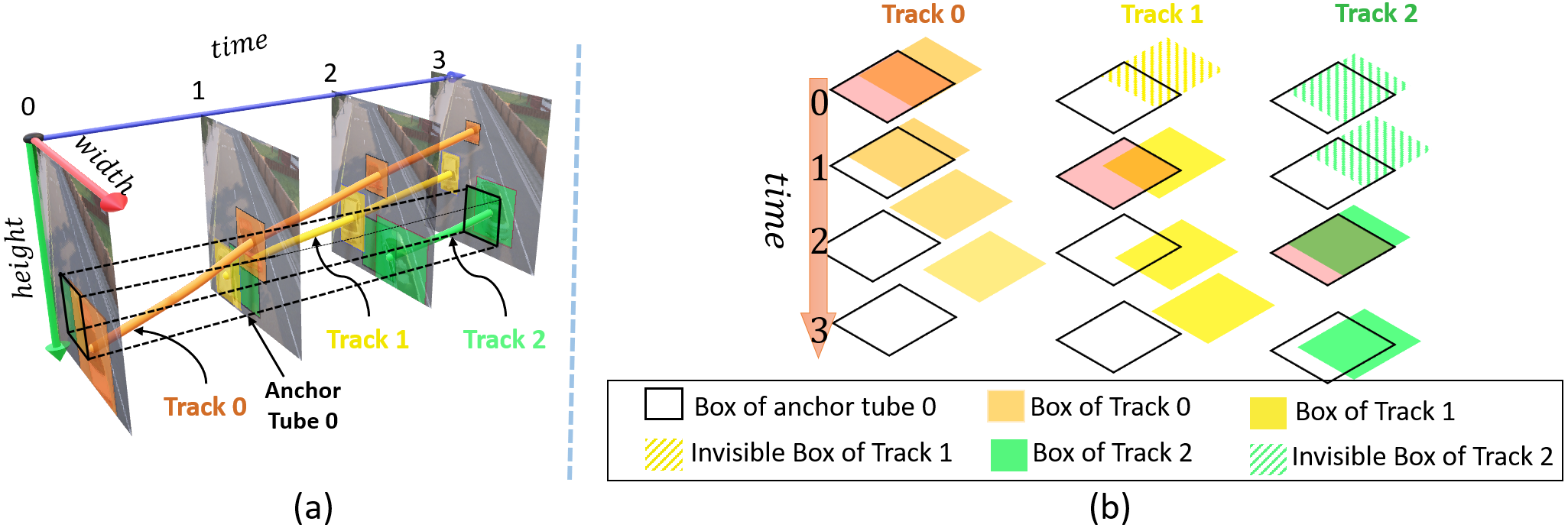}
 \caption{\SJ{(a)} One anchor \SJ{tube} and three tracks are shown to illustrate the search for the best track of the anchor \SJ{tube}. \SJ{(b) is the simplified demonstration of (a)}. Based on the largest overlap between the first \SJ{tube} box and the first {\em visible} track box, Track 2 is selected as the best match. \MS{may be show a real track by connecting the centroid of these boxes?}} 
 %To find the best-matched track of the anchor tunnel-0 , we select the largest overlap ratio between anchor tunnel and tracks. The overlap ratio is calculated by the first visible box of each track. These boxes are labeled with the red border. Apparently, 
 %The Track-2 is the best-matched track of the anchor tunnel-0. }
 \label{fig:matching_strategy}
 %\vspace{-5mm}
 \end{figure}

 Computing anchor \SJ{tubes} for network training can also be interpreted as further data preparation for DMM-Net. We first employ a search strategy to find the best-matched track for each anchor \SJ{tube}\MS{reviewers will be confused with tunnels and tracks, it will be good to use right terminology in order to avoid confusion}, and subsequently encode the \SJ{tube}s by their best-matched tracks. For the former, we specify the overlap between an anchor \SJ{tube} and a track as the overlap ratio between the first visible box of the track and the corresponding box in the anchor \SJ{tube}. A simplified illustration of this concept is presented in Fig.~\ref{fig:matching_strategy} with one anchor \SJ{tube} and three tracks. The anchor \SJ{tube} 0 iterates over all the tracks to find the largest overlap ratio. The first box of Track 1 and the first two boxes of Track 2 are {occluded}. Therefore, the overlap ratio of the anchor box filled red is used for selecting the best-matched track, i.e.~Track 2 for the largest overlap.
 
To encode the anchor \SJ{tubes}, we employ their  best-matched tracks ($\smat B_{t_1:t_2}$) along with the \SJ{classes} ($\smat C_{t_1:t_2}$) and visibility ($\smat V_{t_1:t_2}$). We denote a box of the $i^{\text{th}}$ anchor \SJ{tube} at the $t^{\text{th}}$ frame as $a_{i, t} = (a_{i, t}^{cx}, a_{i, t}^{cy}, a_{i, t}^{w}, a_{i, t}^{h})$, where the superscripts $cx$ and $cy$ indicate the ($x,y$) location of the center, and $w$, $h$ denote the width and height of the box. We use the same notation in the following text as well. Each anchor box is {\color{black}encoded} by the box of its best-matched track as follows:
 %The tracks' localization $B'_{t_1:t_2}$ need to be decoded. We use the anchor \SJ{tube} to decode its matched tracks, as shown in Eq.~\eqref{eq:motion_encode}.
 \begin{equation}
    \left\{\begin{matrix}
    g_{i, t}^w = log(b_{i, t}^w / a_{i, t}^w) \\ 
    g_{i, t}^h = log(b_{i, t}^h / a_{i, t}^h) \\ 
    g_{i, t}^{cx} = (b_{i, t}^{cx} - a_{i, t}^{cx}) / a_{i, t}^w \\
    g_{i, t}^{cy} = (b_{i, t}^{cy} - a_{i, t}^{cy}) / a_{i, t}^h, 
    \end{matrix}\right.
    \label{eq:motion_encode}
 \end{equation}
 where $(b_{i,t}^{cx}, b_{i,t}^{cy}, b_{i,t}^{w}, b_{i,t}^{h})$ describe the box of the best-matched track at the $t^{\text{th}}$ frame, and $(g_{i,t}^{cx}, g_{i,t}^{cy}, g_{i,t}^{w}, g_{i,t}^{h})$ represent the resulting encoded box for the best-matched track. Following this encoding, we replace each original box, \SJ{class}, and visibility by its  newly encoded counterpart.
 %{\color{green}$(b_{i,t}^{cx}, b_{i,t}^{cy}, b_{i,t}^{w}, b_{i,t}^{h})$ is the box of the best-matched track at the $t^{th}$ frame, $(g_{i,t}^{cx}, g_{i,t}^{cy}, g_{i,t}^{w}, g_{i,t}^{h})$ is the result of encoded box.
 %After that, each box, category, and visibility are replaced by the encoded box, category, and visibility of the corresponding best-matched track, respectively. }
 %{\color{red} NOT CLEAR WHAT ARE THE SUPERSCRIPTS. ALSO, WHAT EXACTLY IS `$a$'? IS IT A SCALAR (THEN HOW IS IT REPRESENTING A TRACK?). NEED MORE DETAILS HERE.} 
In the above-mentioned encoding, an anchor \SJ{tube} that has its  best-matched track overlap ratio less than $\delta_o$, is identified as the `background' \SJ{class}. 

 %We fill $\smat B_A$ with the encoded anchor tunnel. 
 %Besides, the $\smat C_A, \smat V_A$ is filled by category and visibility of their best-matched track from $\smat C_{t_1:t_2}$, $\smat V_{t_1:t_2}$.
 %Note that, the encoded anchor tunnel's category is labeled as the background when the overlap ratio of its best-matched tunnel is smaller than $\delta_o$. 
 
 %Finally, the ground truth output of our network is formatted. All the ground tracks' localization, classification, visibility are denoted by $\smat{B}$, $\smat{C}$ and $\smat{V}$ correspondingly.

 \begin{figure}[t]
   \centering
   \includegraphics[width=0.5\linewidth]{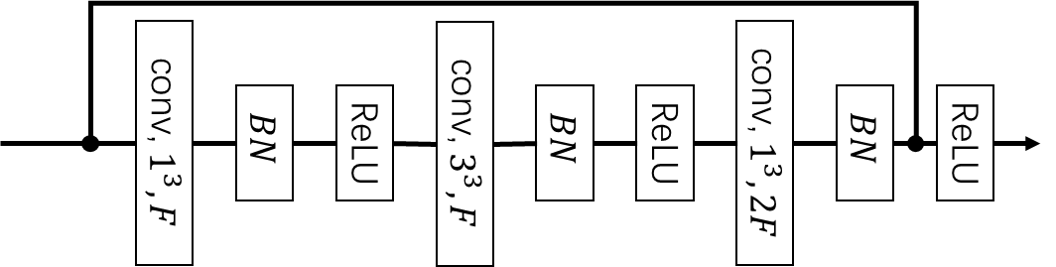}
   \caption{Used ResNeXt block: (conv, x$^3$, F) denote `F' convolutional kernels of size $x \times x \times x$. $BN$ is  Batch Normalization~\cite{ioffe2015batch} and shortcut connection is summation.}
   %\vspace{-4mm}
   \label{fig:ResNeXt}
\end{figure}

%\vspace{-3mm}
 \subsection{Motion Model}
 \label{sec:MM}
 %\vspace{-2mm}
 For motion modeling, we leverage a quadratic model in time.  \MS{what is the motivation of using quadratic, why not linear, or constant velocity as has been used in the past work?}
 One of the outputs of DMM-Net is a tensor of motion parameters $\smat O_M \in \mathbb{R}^{N_T \times 4 \times N_P}$, where $N_P=3$ in our experiments and $N_T$ indicates the number of {\color{black}anchor \SJ{tubes}}.
 We estimate an encoded anchor \SJ{tube} for a track as: 
 \begin{equation}
    \label{eq:motion_model}
    \left\{\begin{matrix}
    \hat{g}_{i,t}^w = p_{11} t^2 + p_{12}t + p_{13} \\ 
    \hat{g}_{i,t}^h = p_{21} t^2 + p_{22}t + p_{23} \\ 
    \hat{g}_{i,t}^{cx} = p_{31} t^2 + p_{32}t + p_{33} \\
    \hat{g}_{i,t}^{cy} = p_{41} t^2 + p_{42}t + p_{43},
    \end{matrix}\right.
 \end{equation}
where $\hat{g}_{i, t}$ indicates an estimated box descriptor of the $i^{\text{th}}$ encoded anchor \SJ{tube} at the $t^{\text{th}}$ frame, \SJ{the superscripts $cx$, $cy$, $w$ and $h$ indicate the center x, center y, height, width of the box respectively,} and $\{p_{11}, \cdots, p_{43}\}$ are the motion parameters\MS{it is not clear what are these motion parameters, width, height, center x, y of each box? then say so clearly.} . Each encoded ground-truth anchor \SJ{tube} can be decoded into a ground truth track by Eq.~\eqref{eq:motion_encode}. We further elaborate on the relationship between the motion parameters and the ground truth tracks in the supplementary material. \SJ{Note that the used motion function is replaceable by any differential function in our technique. We choose quadratic motion modeling considering vehicle tracking as our primary objective.} 
 
%  The ground truth box of the $i^{th}$ encoded anchor tunnel at $t^{th}$ frame also can be described by the motion parameters as shown in Eq.~\eqref{eq:final_model}. 
%  The $(b_{i,t}^{cx}, b_{i,t}^{cy})$ of the track is quadratic function relative to time, while the $b_{i,t}^{w}, b_{i,t}^{h})$ is not quadratic function.
 
%  \begin{equation}
%     \label{eq:final_model}
%     \left\{\begin{matrix}
%     b_{i,t}^{w}  = a_{i,t}^{w}\mathop{exp}({p_{11} t^2 + p_{12}t + p_{13} + \Delta^{w}}) \\ 
%     b_{i,t}^{h}  = a_{i,t}^{h}\mathop{exp}({p_{21} t^2 + p_{22}t + p_{23} + \Delta^{h}}) \\ 
%     b_{i,t}^{cx} = p_{31}a_{i,t}^{w} t^2 + p_{32}a_{i,t}^{w} t + p_{33}a_{i,t}^{w} + a_{i,t}^{cx} + \Delta^{cx} \\ 
%     b_{i,t}^{cy} = p_{41}a_{i,t}^{h} t^2 + p_{42}a_{i,t}^{h} t + p_{43}a_{i,t}^{h} + a_{i,t}^{cy} + \Delta^{cy} 
%     \end{matrix}\right.
%  \end{equation}
 
%  where $\Delta^{\{w, h, cx, cy\}}$ are the localization error of our network output.
 
%\vspace{-4mm}
 \subsection{Architecture}
 %\vspace{-2mm}
 As shown in Fig.~\ref{fig:pipeline}, our network comprises a Feature Extractor and three sub-network. The network simultaneously  processes $N_F$ video frames for which features are first extracted. The feature maps  of six intermediate layers of the Feature Extractor are fed to the sub-networks. These networks predict tensors containing  motion parameters ($\smat O_M \in \mathbb{R}^{N_T\times N_P \times 4}$), object \SJ{classes} ($\smat O_C \in \mathbb{R}^{N_T \times N_C}$), and visibility information ($\smat O_V \in \mathbb{R}^{N_F \times N_T \times 2}$).

 %(namely, motion subnet, classifier subnet, and visibility subnet) for estimating the motion parameters, categories, and visibilities. 
 %As shown in Fig.~\ref{fig:pipeline}, $N_F$ video frames are firstly processed by the feature extractor. After that, we select 6 intermediate feature maps and input these feature maps into these three subnets. These subnets separately ouput 3 estimated matrices, motion parameters matrice ($\smat O_M \in \mathbb{R}^{N_T\times N_P \times 4}$), category matrice ($\smat O_C \in \mathbb{R}^{N_T \times N_C}$), and visibility matrice ($\smat O_V \in \mathbb{R}^{N_F \times N_T \times 2}$).

 %\vspace{-3mm}
 \noindent {\bf Feature Extractor:}
 We construct our Feature Extractor based on the ResNeXt blocks of the 3D ResNet~\cite{Hara2018}\blfootnote{foot:notation}{Notation are adopted from the original work.}. The architectural details of the blocks are provided in Fig.~\ref{fig:ResNeXt}. The blocks accept `$F$' channel input that allows us to simultaneously model spatio-temporal features in $N_F$ frames. We enhance the 3D ResNet architecture by removing the \textit{fully-connected} and \textit{softmax} layers of the network and appending two extra convolutional groups denoted as Conv6\_x and Conv7\_x.
 We perform this enhancement because the additional convolutional layers are still able to encode further higher level spatio-temporal features of the frames.  
 Details on the convolutional groups used by the Feature Extractor are provided in Table~\ref{tab:feature_extractor}. Here, Conv1\_x (abbreviated as C1\_x) contains 64 convolutional kernels of size $7 \times 7 \times 7$, stride $ 1 \times 2 \times 2$. A $3\times 3 \times 3$ max-pooling layer with stride 2 is inserted before Conv2\_x for down sampling. Each convolutional layer is followed by batch normalization~\cite{He2016a} and ReLU~\cite{Nair2010b}. Spatio-temporal down-sampling is performed by Conv3\_1, Conv4\_1 and Conv5\_1 with a stride of 2. %({\color{red} are we using the terms 'group', 'block' and 'layer' correctly here? {\color{green}(yes, it's right. these notations are consistent with the paper of 3D ResNet.)}})

 \begin{table}[t]
   \begin{center}
      \small
        \caption{Convolutional groups of the Feature Extractor. Conv is abbreviated as `C', `F' is the number of feature maps, as in Fig.~\ref{fig:ResNeXt}, and `N' is the number of bottleneck blocks in each layer. %{\color{red} Why we have x in C2\_x? to shows is a group and not a layer?}
   %{\color{green} (This type of representation comes from the paper of 3D ResNet~\cite{Hara2018}. The ``x'' is the index of the basic block. If the ``x'' is not specified, the ``C1\_x'' represents a group. The ``C1\_1'' is the first block in C1\_x group.)}
      }
      \label{tab:feature_extractor}

      \tabcolsep=3pt
      \begin{tabular}{|@{}c|c|c|c|c|c|c|c|c|c|c|c|c|@{}|}
      \hline
      % \multicolumn{9}{|c}{3D Resnet}                   & \multicolumn{4}{|c|}{Extension} \\ 
      % \hline\hline
      \multicolumn{1}{|c}{C1\_x}                       & \multicolumn{2}{|c}{C2\_x} & \multicolumn{2}{|c}{C3\_x} & \multicolumn{2}{|c}{C4\_x} & \multicolumn{2}{|c}{C5\_x} & \multicolumn{2}{|c}{C6\_x} & \multicolumn{2}{|c|}{C7\_x} \\ 
      \hline
      \multirow{2}{*}{conv, $7^3$, 64} & F            & N          & F            & N          & F            & N          & F            & N          & F            & N          & F            & N          \\ 
      \cline{2-13}\cline{2-13}
                                                      & 64           & 3          & 128          & 4          & 256          & 23         & 512          & 3          & 32           & 3          & 32           & 3          \\ \hline
      \end{tabular}
   \end{center}
  % %\vspace{-3mm}
   \end{table}

%\vspace{-2mm}
 \noindent {\bf Sub-networks:} The DMM-Net has three sub-networks to compute motion parameters, object \SJ{classes}, and their visibility.
These networks use the outputs of Conv2\_x to Conv7\_x groups of the Feature Extractor.
Each sub-network processes the input with six convolutional layers.
The architectural details of all three sub-networks are summarized in Table~\ref{tab:motion_branch}.
The kernel shape, padding size and stride are the same for each network, whereas the employed numbers of kernels are different.
We fix the temporal kernel size for each network to  $\{8, 4, 2, 1, 1, 1\}$. 
Denoting the numbers of anchor \SJ{tubes} defined for the six input feature maps of a sub-network by $\sset K$, we let $\sset K = \{10, 8, 8, 5, 4, 4\}$ in our experiments.   For each sub-network, we concatenate the output feature maps of each convolutional layer. 
We reshape the concatenated feature according to the  dimensions mentioned in the last row of the table.
These outputs are subsequently used to compute the network loss. 

% ``Output'' in Tab.~\ref{tab:motion_branch}.

%Each subnet has six convolutional layers to process the input feature maps separately. 
%As shown in Tab.~\ref{tab:motion_branch}, the convolutional layers of each branch share the same padding size, and stride size, while the kernel number and kernel size are different. 
%$\sset K$ is a set of anchor tunnel numbers at each position of the feature map. 
%In our experiment, $\sset K = \{10, 8, 8, 5, 4, 4\}$. 
%Therefore, the kernel number of motion branch are $\{40N_P, 32N_P, 32N_P, 20N_P, 16N_P, 16N_P\}$. 
%Besides, in our experiment, we use $\{8, 4, 2, 1, 1, 1\}$ as the temporal kernel size for each subnet. 
 %For each subnet, we concatenate the result feature map of each convolutional layer. 
 %After that, we reshape the concatenated result according to the row ``Output'' in Tab.~\ref{tab:motion_branch}.

%, namely, 1) the \textit{Motion Subnet}, 2) the \textit{Classifier Subnet}, and 3) the \textit{Visibility Subnet}. 
% They accept six intermediate feature maps from the \textit{Feature Extractor} and are responsible for the estimation of motion parameters, categories, and visibilities. 
% As shown in Fig.~\ref{fig:pipeline}, these six feature maps are the output of Conv2\_x to Conv7\_x respectively. 
 
 \begin{table}[t]
 \begin{center}
    \small
     \caption{Details of  the {Motion Subnet} (M.S.),  {Classifier Subnet} (C.S.), and {Visibility Subnet} (V.S.): 
   K.S., P.S., S.S., and N.K. respectively denote the kernel shape, padding size, stride size, and the number of kernels. 
   The row `Output' reports output tensor shape. 
   $\sset K=\{10, 8, 8, 5, 4, 4\}$ is the number of anchor \SJ{tubes} for each of the six input feature maps for each sub-network. }
   \label{tab:motion_branch}

    \begin{tabular}{l|c|c|c|}
    \cline{2-4}
                                    & M.S.                      & C.S.                   & V.S.                      \\ \hline
    \multicolumn{1}{|l|}{K.S.}  &\multicolumn{3}{|c|}{$\{8, 4, 2, 1, 1, 1\}\times 3 \times 3$}  \\ \hline
    \multicolumn{1}{|l|}{P.S.}   & \multicolumn{3}{|c|}{$0 \times 1 \times 1$}     \\ \hline
    \multicolumn{1}{|l|}{S.S.}    & \multicolumn{3}{|c|}{$1 \times 1 \times 1$}     \\ \hline
    \multicolumn{1}{|l|}{N.K.}    & $4\sset{K}N_P$                 & $\sset{K}N_C$               & $2\sset{K}$                    \\ \hline
    \multicolumn{1}{|l|}{Output} & $N_T \times N_P \times 4$ & $N_T \times N_C$       & $N_F \times N_T \times 2$ \\ \hline
    \end{tabular}
 \end{center}
 %\vspace{-5mm}
 \end{table}
 
%  Each subnet has six convolutional layers to process the input feature maps separately. 
%  As shown in Tab.~\ref{tab:motion_branch}, the convolutional layers of each branch share the same padding size, and stride size, while the kernel number and kernel size are different. 
%  $\sset K$ is a set of anchor tunnel numbers at each position of the feature map. 
%  In our experiment, $\sset K = \{10, 8, 8, 5, 4, 4\}$. 
%  Therefore, the kernel number of motion branch are $\{40N_P, 32N_P, 32N_P, 20N_P, 16N_P, 16N_P\}$. 
%  Besides, in our experiment, we use $\{8, 4, 2, 1, 1, 1\}$ as the temporal kernel size for each subnet. 
%  For each subnet, we concatenate the result feature map of each convolutional layer. 
%  After that, we reshape the concatenated result according to the row ``Output'' in Tab.~\ref{tab:motion_branch}.

 %\vspace{-2mm}
\noindent {\bf Network Loss:} Based on the architecture, the overall loss of DMM-Net is defined as a weighted sum of three sub-losses, including the motion loss ($\sfun L_M$), the classification loss ($\sfun L_C$) and the visibility loss ($\sfun L_V$), given as:
 \begin{equation}
    \begin{aligned}
       \label{eq:loss}
       \sfun L = \frac{1}{N} (\alpha \sfun L_M + \beta  \sfun L_C + \sfun L_V),
    \end{aligned}    
 \end{equation}
where $N$ is the number of \textit{positive} anchor \SJ{tubes}, and $\alpha$, $\beta$ are both hyper-parameters. The positive anchor \SJ{tubes} exclude the background \SJ{tubes}.

%For DMM-Net, each anchor tunnel has a best-matched ground truth track. In other words, $\sum_i x_{i,j,t}^p \geq 1$ for any $t$. 
% The overall loss function is a weighted sum of three sub-losses, including the localization loss ($\sfun L_M$), the classification loss ($\sfun L_C$) and the visibility loss ($\sfun L_V$), as follows:

 % \begin{equation}
 % \begin{aligned}
 %    \label{eq:loss}
 %    &\sfun L(x, c, l, g, v, e) = \\
 %    &\frac{1}{N} (\alpha \sfun L_M(x, l, g) + \beta  \sfun L_C(x, c) + \sfun L_V(x, v, e))
 % \end{aligned}    
 % \end{equation}

 % \begin{equation}
 % \begin{aligned}
 %    \label{eq:loss}
 %    &\sfun L(x, \smat B, \smat C, \smat V, \smat O_M, \smat O_C, \smat O_V) = \frac{1}{N} \\
 %    &(\alpha \sfun L_M(x, \smat B, \smat O_M) + \beta  \sfun L_C(x, \smat C, \smat O_C) + \sfun L_V(x, \smat V, \smat O_V))
 % \end{aligned}    
 % \end{equation}

 %Where $N$ is the number of positive anchor \SJ{tubes}, both $\alpha$ and $\beta$ are balance factors. We ignore the case when $N=0$. 

 The motion loss $\sfun L_{M}$ is defined as the sum of \textit{Smooth}-L1 losses between the ground truth encoded tracks $g_{i,t}$ and their predicted encoded tracks $\hat{g}_{i,t}$, formally:
 \begin{equation}
    \label{eq:loc-loss}
   %  \resizebox{0.43\textwidth}{!}{$
    \sfun L_{M} = \sum_{t=0}^{N_F-1}\sum_{i\in \mathop{Pos}} \sum_{m \in \{cx, cy, h, w\}} ||g_{i, t}^m - \hat{g}_{i,t}^m||_1,
   %  $s}
 \end{equation}
 where $Pos$ is the set of positive encoded anchor \SJ{tube} indices. %, $g_{i,t}$ is the box of the $i^{th}$ encoded anchor tunnel at $t$-th frame,
 %and $\hat{g}_{i, t}^m$ are the corresponding estimated box.
 We compute $\hat{g}_{i,t}^m$  using Eq.~\eqref{eq:motion_model} discussed in Section~\ref{sec:MM}.
 
 %  where $\smat{B}$ is the location ground truth, $\smat O_M \in \mathbb{R}^{N_T\times 4\times 3}$ is the predicted motion parameter matrix
 
 % \begin{equation}
 %    \label{eq:motion}
 %    \left[\begin{matrix}
 %    \hat{g}_{i,t}^{cx} \\
 %    \hat{g}_{i,t}^{cy} \\
 %    \hat{g}_{i,t}^{h} \\
 %    \hat{g}_{i,t}^{w} 
 %    \end{matrix}\right] = \hat{\smat{P}}_i \times 
 %    \left[\begin{matrix}
 %    t^2 \\
 %    t \\
 %    1
 %    \end{matrix}\right]
 % \end{equation}

For the classification loss $\sfun L_{C}$, we employ the hard negative mining strategy~\cite{Liu2016} to balance the positive and negative anchor \SJ{tubes}, where the negative \SJ{tubes} correspond to the background.
We denote $x_{i,j,t}^p \in \{1, 0\}$ to be an indicator for matching the $i^{\text{th}}$ anchor \SJ{tube} and the $j^{\text{th}}$ ground-truth track of \SJ{class} `$p$' at the $t^{\text{th}}$ frame. 
Each anchor \SJ{tube} at least has one best-matched ground-truth track, implying $\sum_i x_{i,j,t}^p \geq 1$ for any $t$.
%The classification loss accumulates the $\mathop{softmax}$ scores over multiple categories confidence $(c)$ for different frames, as shown in Eq.~\eqref{eq:confidence}.
The classification loss of the DMM-Net is defined as: 

   \begin{equation}\label{eq:confidence}
          \sfun L_{C}= \sum_{t=0}^{N_F}\left(
          -\sum_{i\in \mathop{Pos}} x_{i,j,t}^p \mathop{log}(\hat{c}_{i,t}^p) - 
          \sum_{i\in \mathop{Neg}} \mathop{log}(\hat{c}_{i,t}^0)
          \right),
    \end{equation}
where $\hat{c}_{i,t}^p = \frac{\mathop{exp}c_{i,t}^p}{\sum_p \mathop{exp} c_{i,t}^p}$ such that $c_{i, t}^p$ is the  predicted  confidence of the object being for class $p: p \in \{1, \cdots, N_C\}$, $Neg$ is the set of negative encoded anchor \SJ{tube} indices. 
We also consider the visibility estimation task fro  classification viewpoint and \SJ{classify} each positive box into invisible `0' or visible `1' box. Based on that, the loss is computed as: 
\begin{equation}
    \label{eq:visibility}
    \sfun L_{V} = \sum_{t=0}^{N_F}\left(
    -\sum_{i \in \mathop{Pos}} y_{i,j,t}^q \mathop{log}(\hat{v}_{i,t}^q)
    \right),
 \end{equation}
where $\hat{v}_{i,t}^q = \frac{\mathop{exp}v_{i,t}^q}{\sum_q\mathop{exp}v_{i,t}^q}$, and $y_{i,j,t}^q$ is an indicator for matching the $i^{\text{th}}$ anchor \SJ{tube} and the $j^{\text{th}}$ ground-truth track of visibility `$q$' at the $t^{\text{th}}$ frame, such that  $q \in \{0, 1\}$.
 
 % \subsection{Motion Model}
 % %their motions pattern ary different in the image. 
 % Most of objects' motion pattern in the 3D world coordinate is rigid. As our network takes frames within a short time interval, these motion patterns can be approximated to uniform linear motion in the 3D world coordinate. Based on the transformation between world coordinate and image coordinate, we design a motion function which can well fit the motion pattern in the image coordinate.
 
 % We assume the motion pattern of object in 3D world coordinate is uniform linear. $P_W \in \mathbb{R}^{3}$ is a point in the world coordinate. The motion equation of $P_W$ with respect to time $t$ as follows:
 % \begin{equation}
 %     \label{eq:motion_world}
 %     P_W = P_1t + P_2
 % \end{equation}
 % where, $P_1 \in \mathbb{R}^{3}, P_2 \in \mathbb{R}^3$ are the motion parameters.
 
 % The camera parameters matrix is $C_P \in \mathbb{R}^{3\times 4}$ defined as follows:
 % $$
 % C_P = \left[\begin{matrix}
 %         r_{11} & r_{12} & r_{13} & r_{14} \\
 %         r_{21} & r_{22} & r_{23} & r_{24} \\
 %         0 & 0 & 0 & 1 \\
 %     \end{matrix}\right]
 % $$
 
 % Therefore, the projection point $P_I$ of $P_W$ in the image coordinate can be:
 % $$
 % P_I = C_P P_W
 % $$
 
 \begin{figure}[t]
   \centering
   \includegraphics[width=0.7\linewidth]{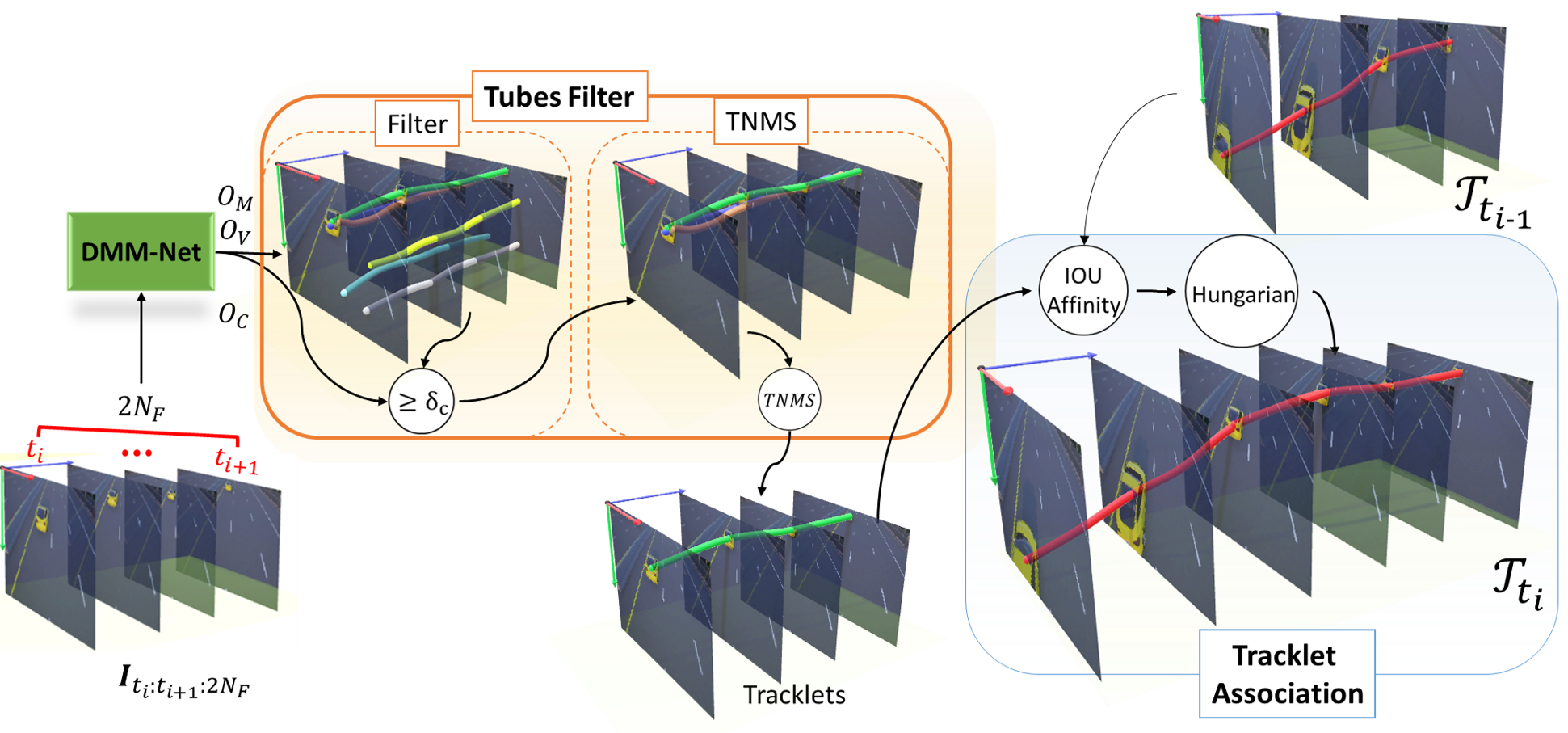}
   \caption{Deployment of DMM-Net: For the $2N_F$ frames $\smat I_{t_i:t_{i+1}:2N_F}$ from $t_i$ to $t_{i+1}$, the trained DMM-Net selects $N_F$ frames as its input and outputs predicted \SJ{tubes} encoded by object motion parameter matrix $(\smat O_M)$, \SJ{classification} matrix $(\smat O_C)$ and visibility matrix $(\smat O_V)$. These matrices are filtered and the track set $\sset T_{t_i}$ is updated by associating the filtered \SJ{tracklets} by their IOU with the previous track set $\sset T_{t_{i-1}}$.}
   \label{fig:tracker}
   %\vspace{-5mm}
\end{figure}

%\vspace{-3mm}
 \subsection{Deployment}
 The trained DMM-Net is readily deployable for tracking (Fig.~\ref{fig:tracker}). The overall tracker processes $2N_F$  frames $\smat I_{t_i:t_{i+1}:2N_F}$, where the DMM-Net first selects $N_F$ consecutive frames and outputs the predicted tubes encoded with object motion parameters, \SJ{classes}  and visibility using Eq.~\eqref{eq:motion_model}. The \SJ{tube}s are decoded with Eq.~\eqref{eq:motion_encode}. A filtration  is then done to remove the undesired \SJ{tubes} and we get tracklets (details to follow).
 We compute updated  trajectory set $\sset{T}_{t_i}$ by associating the \SJ{tracklets} with the previous trajectory set $\sset T_{t_{i-1}}$ using their IOUs.

To filter, we first remove \SJ{tubes} with confidence lower than a threshold $\delta_c$. We subsequently perform a \SJ{Tube} None Maximum Suppression (TNMS).
To that end, we cluster the detected \SJ{tubes} of the same category into multiple groups by their IOUs with a threshold $\delta_{nms}$.
Then, we only keep one \SJ{tube} for each group that has the maximum confidence of being positive. \SJ{The kept tubes, namely ``tracklets'', are employed to update trajectory set.}
%To remove unreasonable tunnels, we propose a Tunnel Filter. In detail, we first remove the tunnels with lower confidence than $\delta_c$. After that, the Tunnel None Maximum Suppression (TNMS) is employed for further filter these overlapped detected tunnels, as shown in Fig.~\ref{fig:tracker}.
%Specifically, for the TNMS, we first cluster detected tunnels of the same category into multiple groups by IOU of tunnels with a threshold $\delta_{nms}$.
%Then, for each group, we only keep one tunnel with the maximum confidence of being positive.

We initialize our track set $\sset T_{t_i}$ with as many trajectories as the number of \SJ{tracklets}. %The tracks in our settings can be seen as a set of \SJ{tracklets}. 
The track set is updated from $t_i^{\text{th}}$ to $t_{i+1}^{\text{th}}$ stamp using the Hungarian algorithm~\cite{Munkres1957} applied to an IOU matrix $\smat \Psi \in \mathbb{R}^{n'_{t_{i-1}}\times n_{t_i}}$,
where $n'_{t_{i-1}}$ is the number of element in the previous track set, and $n_{t_i}$ is the number of \SJ{tracklets}. 
Notice that we perform association of tracklets and not the individual objects. The object association remains implicit and is done inside the DMM-Net which lets us define the \SJ{tube}s. The association of tracklets, defined across multiple frames, leads to significant computational gain. This is one of the core reasons of the orders of magnitude gain in the tracking speed of our technique over existing methods, as will be clear in Section~\ref{sec:e}.

\begin{figure}[t]
   \centering
   \includegraphics[width=0.5\linewidth]{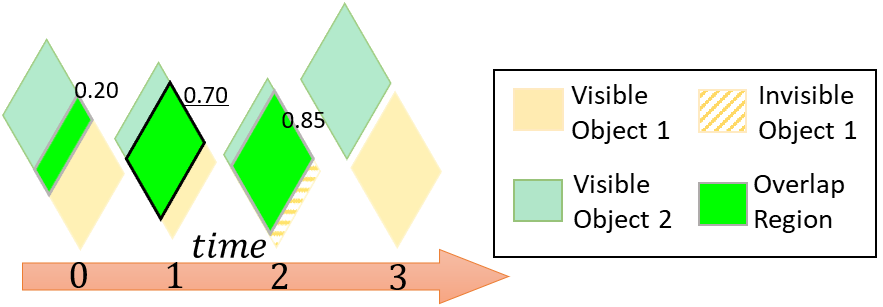}
   \caption{Example of calculating \SJ{Tube} IOU. There are two \SJ{tracklets} (yellow, cyan). Their intersection is green. The \SJ{tracklet} IOU is the maximum IOU of the visible box pair.  Although  IOU at $t=2$ is the largest, the lined yellow object is invisible, hence, we select the second largest IOU at $t=1$ as the \SJ{tracklet} IOU.}
   \label{fig:nms}
   %\vspace{-5mm}
\end{figure}

To form the matrix $\smat \Psi$, we must compute IOU between the existing \SJ{tracklets} and the new \SJ{tracklets}.
Fig.~\ref{fig:nms} shows the procedure adopted for calculating the IOU between two \SJ{tracklets} with  a simplified example.  
Overall, our tracker is an on-line technique in the sense that it does not use future frames to predict object trajectories.
Concurrently, DMM-Net can perform tracking in real-time  that makes it a highly desirable technique for practical applications.

 \section{Experiments}\label{sec:e}
 %\vspace{-2mm}
 We evaluate our technique using the proposed Omni-MOT dataset and the popular UA-DETRAC challenge~\cite{Wen2015a}.
 The former is to demonstrate both the effectiveness of our network for MOT and trainability of deep models for MOT with Omni-MOT. 
 We mainly focus on `vehicle' tracking in this work. The proposed dataset is also for the vehicle tracking problem.
 Whereas  DMM-Net is trained here for vehicle tracking, it is possible to train it for e.g.~pedestrian tracking. %However, due to the complex (detector + tracker) modeling performed by our network, it requires much larger amount of training data that commensurate with the modeling task complexity. 
 %Existing pedestrian tracking challenges e.g.~MOT15, 17 provide relatively less training data. Moreover, the evaluation servers compute results that are influenced by off-the-shelf detectors that automatically account for the covariate shift between the training and testing data. These issues do not allow a fair benchmarking of our technique on these online challenges. Hence, we focus on the vehicle tracking problem in our experiments, for which our dataset can further help in fair evaluation.    
 
 %to acquire reasonable performance with DMM-Net architecture, larger datasets with accurate annotations are required due to the complex (detector + tracker) modelling performed by our network. The current datasets for pedestrian tracking e.g.~MOT15, 17, do not fulfill this requirement. Hence, both our dataset and experiments focus on vehicle tracking.   

 %\SJ{Notice that our network cannot achieve a good result on MOT17 as there are a big gap between testing dataset and training dataset.}
 
 %In this section, we evaluate the proposed approach on two datasets, namely OMOTD (our proposed) and UA-DETRAC~\cite{Wen2015a}. 
 %The UA-DETRAC dataset is on-line challenge where a hosting server evaluates the submitted methods. 
 %We first report the result of the implementation detail of our proposed technique, followed by these two datasets. 
 %The performance of our approach of UA-DETRAC challenge and its comparison to other techniques currently on the leader board is also presented.
 
   % Please add the following required packages to your document preamble:
 % \usepackage{multirow}

%\vspace{-3mm}
 \noindent{\bf Implementation Details:}
 We implement the DMM-Net using  Pytorch~\cite{Paszke2017} and train it using NVIDIA GeForce GTX Titan GPU. 
 The hyper-parameter values are selected with cross-validation to maximize the MOTA metric on the validation set of  UA-DETRAC. The chosen values of the hyperparameters are as follows. 
 Batch size $B = 8$, number of training epochs per model $= 20$, number of input frames $= 16$, input frame size $= 168 \times 168$, 
 We use Adam Optimizer~\cite{kingma2015adam} for training. Other hyper-parameter values are $\delta_{c}=0.4, \delta_{nms}=0.3,$ and $\delta_o=0.5$.

% \begin{figure*}[h]
%    \centering
%    \includegraphics[width=6.5in]{imgs/amot_result.png}
%    \caption{Tracking result from OMOTD.}
%    \label{fig:omot_result}
%    %\vspace{-3mm}
% \end{figure*}

%\vspace{-3mm}
\noindent{\bf Omni-MOT dataset: }
 To demonstrated the efficacy of DMM-Net and trainability of deep models on our  dataset, we select five scenes from the Omni-MOT and perform tracking.  
 The scenes are chosen with Easy camera viewpoint with clear weather conditions. The selected  scenes are from Town 02 (with 50 vehicles) that are indexed 0, 1, 5, 7 and 8 in the dataset. In this experiment, we train and test DMM-Net using the train and test partitions of the selected scene, as provided by Omni-MOT. Our training required 59.2 hours for 22 epochs.  

  We use both CLEAR MOT~\cite{Bernardin2008}, and MT/ML~\cite{Ristani2016} metrics and summarize the results in Table~\ref{tab:result-omot}. For the definitions of  metrics we refer to the original works. We refer to  \cite{Leal-Taixe2015} for the details on the comprehensive metrics MOTA and MOTAP.
 The table provides metric values for both training and test partitions. The variation between these values is a clear indication that despite the Easy camera view, the dataset provides reasonable challenges for the MOT task. We provide results with additional view points in the supplementary material to further put the reported values into better  perspective.
 
 \begin{table*}[t]
   \begin{center}
         \caption{Results on Omni-MOT: The symbol $\uparrow$ indicates higher values are better, and $\downarrow$ implies lower values are favored. \MS{Why some of the metrics are not common in both Tabels 4 and 5; reviewers will ask?} }
         \label{tab:result-omot}
         %\vspace{-5mm}
      \resizebox{\textwidth}{!}{
      \begin{tabular}{llccccccccccccccc}
      \hline
      \textbf{Type}          & \textbf{Camera}   & \textbf{IDF1}$\uparrow$ & \textbf{IDP}$\uparrow$ & \textbf{IDR}$\uparrow$ & \textbf{Rcll}$\uparrow$ & \textbf{Prcn}$\uparrow$ & \textbf{GT}$\uparrow$ & \textbf{MT}$\uparrow$ & \textbf{PT}$\uparrow$ & \textbf{ML}$\downarrow$ & \textbf{FP}$\downarrow$ & \textbf{FN}$\downarrow$ & \textbf{IDs}$\downarrow$ & \textbf{FM}$\downarrow$ & \textbf{MOTA}$\uparrow$ & \textbf{MOTP}$\uparrow$ \\ 
      \hline
      \multirow{5}{*}{Test}  & Camera\_0 & 72.2\%        & 69.3\%       & 75.4\%       & 90.6\%        & 83.2\%        & 41          & 38          & 3           & 0           & 1762        & 911         & 18           & 98          & 72.1\%        & 79.3\%        \\
                             & Camera\_1 & 59.3\%        & 56.2\%       & 62.7\%       & 81.9\%        & 73.4\%        & 35          & 19          & 15          & 1           & 1199        & 730         & 12           & 61          & 52.0\%        & 75.5\%        \\
                             & Camera\_5 & 40.2\%        & 61.3\%       & 29.9\%       & 34.7\%        & 71.1\%        & 44          & 9           & 20          & 15          & 2779        & 12855       & 38           & 80          & 20.4\%        & 76.1\%        \\
                             & Camera\_7 & 77.8\%        & 79.6\%       & 76.1\%       & 88.3\%        & 92.4\%        & 37          & 29          & 8           & 0           & 1518        & 2457        & 20           & 99          & 80.9\%        & 80.3\%        \\
                             & Camera\_8 & 30.5\%        & 47.3\%       & 22.5\%       & 33.5\%        & 70.7\%        & 38          & 0           & 28          & 10          & 1789        & 8548        & 36           & 110         & 19.3\%        & 70.3\%        \\\hline 
      \multirow{5}{*}{Train} & Camera\_0 & 59.2\%        & 58.6\%       & 59.7\%       & 87.7\%        & 86.0\%        & 48          & 40          & 8           & 0           & 2517        & 2174        & 43           & 172         & 73.2\%        & 79.5\%        \\
                             & Camera\_1 & 47.6\%        & 44.2\%       & 51.6\%       & 80.0\%        & 68.6\%        & 46          & 29          & 16          & 1           & 3681        & 2005        & 50           & 137         & 42.9\%        & 76.9\%        \\
                             & Camera\_5 & 45.0\%        & 52.7\%       & 39.2\%       & 57.0\%        & 76.7\%        & 49          & 6           & 37          & 6           & 4844        & 12000       & 122          & 234         & 39.2\%        & 77.3\%        \\
                             & Camera\_7 & 74.7\%        & 71.9\%       & 77.8\%       & 90.1\%        & 83.3\%        & 43          & 31          & 12          & 0           & 1804        & 985         & 16           & 91          & 71.8\%        & 79.7\%        \\
                             & Camera\_8 & 41.7\%        & 52.2\%       & 34.7\%       & 44.1\%        & 66.3\%        & 43          & 0           & 41          & 2           & 1884        & 4705        & 18           & 117         & 21.4\%        & 68.3\%        \\ \hline 
      \multicolumn{2}{c}{Average}            & 55.5\%        & 61.1\%       & 50.9\%       & 66.5\%        & 79.8\%        & 424         & 201         & 188         & 35          & 23777       & 47370       & 373          & 1199        & 49.4\%        & 77.8\%        \\ \hline
      \end{tabular} }
   \end{center}
   %\vspace{-3mm}
\end{table*}

 \begin{figure*}[t]
   \centering
   \includegraphics[width=0.75\textwidth] {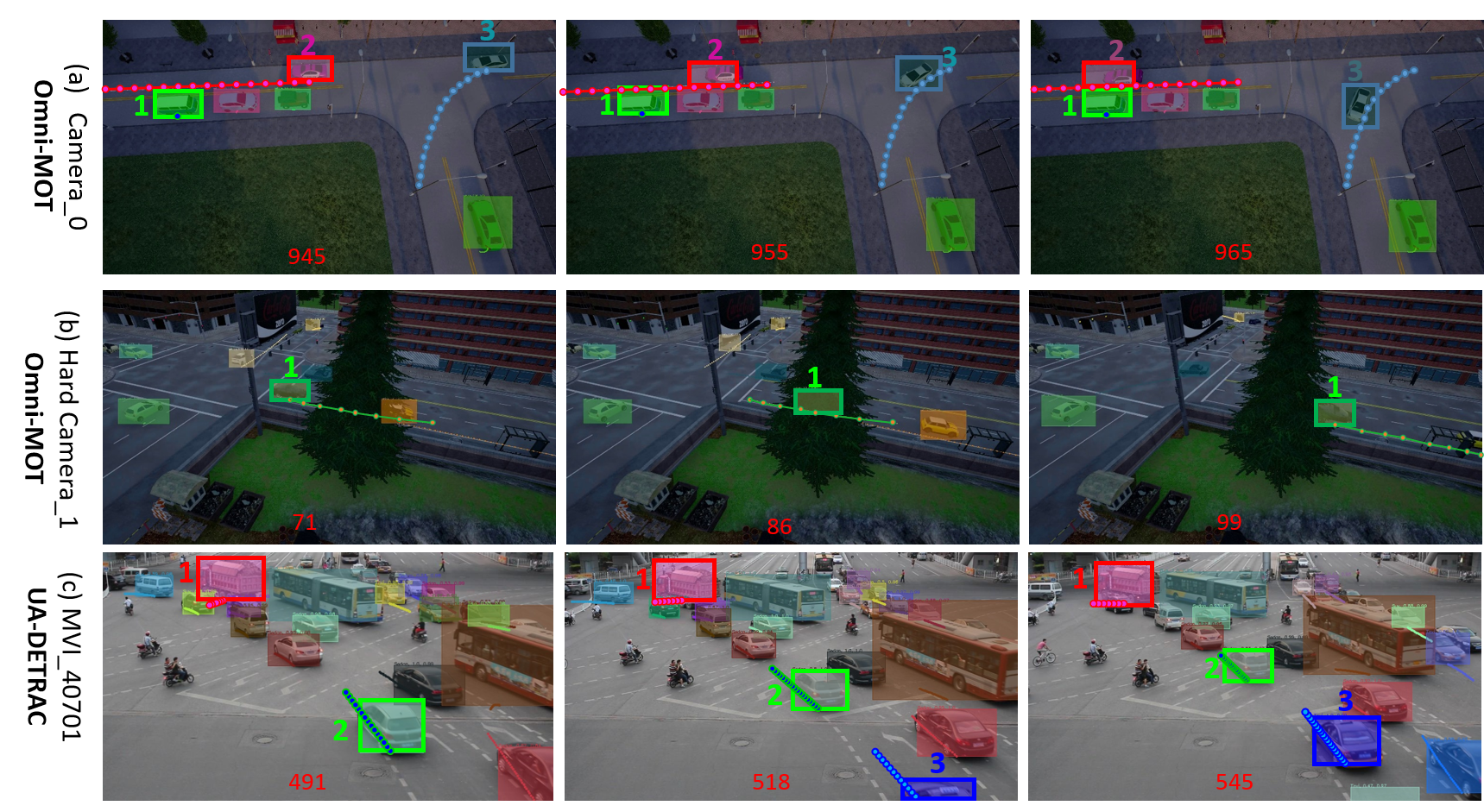}
   \caption{Tracking illustration of DMM-Net on Omni-MOT (first and second row) and  UA-DETRAC (third row). 
   The mentioned IDs are for reference in the text only. \MS{This does not seem that impressive, there are only three objects. May be show examples with more objects. Ana may be show tracks in one image instead of showing all frames.}}
   \label{fig:ua_omot_examples}
   %\vspace{-5mm}
\end{figure*}

 An illustration of tracking for Camera$\_0$ is also provided in Fig.~\ref{fig:ua_omot_examples} (top). Our technique is able to easily track vehicles that are stationary, (e.g.~1), moving along a straight path (e.g.~2), and moving along a curved path (e.g.~3), which justifies the selection of our motion model. \SJ{Additionally, our tracker has the power to deal with even full occlusions between the frames. We show such a case for a hard camera view point in the middle row of  Fig.~\ref{fig:ua_omot_examples}. In this case, vehicle-1 is totally occluded at the $86^{th}$ frame which can be detected and tracked by our tracker. At the $99^{th}$ frame, vehicle-1 is partially occluded, but the tracklets association performs correctly. Note that, our tracker does not need to be evaluated with (detector-based) UA-DETRAC metrics~\cite{Wen2015a} as the Omni-MOT dataset provides ground truth detections without using off-the-shelf detectors. }
 %The tracking results of our approach are also illustrated with the help of examples in Fig.~\ref{fig:ua_omot_examples}.
 %The mentioned object is highlighted, and their identification numbers are for reference in the text. 
 %In the first row of Fig.~\ref{fig:ua_omot_examples}, the network can locate targets of various movement patterns, for instance,
 %identity-1 with static motion, identity-2 with linear motion, and identity-3 with bending motion.
%  1) static motion (i.e., object 1, 2) linear motion (i.e., object 2), and 3) bending motion (i.e., object 3).
%  Moreover, the tracker can connect different motion patterns. For example, in the ``Easy\_Camera\_8'', the network predicts the motion parameter of object 1 both from frame 3 to frame 34th and from frame 30 to frame 61.
%  After that, the tracker connects these two motion parameters based on the overlapped frame (30 to 34).

%  \begin{figure}[t]
 
%     \centering
%     \includegraphics[width=2.5in]{imgs/amot_loss.png}
%     \caption{The loss result of OMOT dataset.}
%     \label{fig:omot_loss}
%  \end{figure}

%  \begin{figure*}[t]
%   \centering
%   \sidebysidecaption{0.19\linewidth}{0.8\linewidth}
%   {%
%   \caption{ {\color{red} Use full frames for 965 and 545. Reduce size for space.}Tracking illustration of DMM-Net on (a) Omni-MOT and (b) UA-DETRAC. 
%   The mentioned IDs are for reference in the text only.}
%   \label{fig:ua_omot_examples}
%   }{
%   \includegraphics[width=0.9\linewidth]{imgs/ua_omot_result.png}
%   }
% \end{figure*}

\noindent{\bf UA-DETRAC: }
 %\vspace{-1mm}
 The UA-DETRAC challenge \cite{Wen2015a} is arguably the most widely used large-scale benchmark for MOT of vehicles. It comprises 100 videos @ 25 fps, recorded in 24 locations with frame size $540\times960$. For this challenge, results are computed by a remote server using CLEAR MOT and MT/ML metrics with Precision-Recall curve of the detection.  We refer to \cite{Wen2015a} for further details on the metrics.
 In Table~\ref{tab:result-ua}, we show our results on the UA-DETRAC challenge. The pre-fix PR for the metrics indicates  the use of Precision-Recall curve. It can be seen that DMM-Net  achieves excellent MOTA score, which is widely considered as the most comprehensive metric for MOT. We also provide results for DMM-Net+ which augments the UA-DETRAC training set with a subset of Omni-MOT.  This subset contained 8 random videos from Omni-MOT. We can see an overall performance gain with this augmentation, exemplifying the utility of Omni-MOT in data augmentation. 
 
 \begin{table*}[t]
   \begin{center}
      \small
       \caption{Results on UA-DETRAC:
    `T.S.' is  tracker speed (fps), `D.S.' is  detector speed  (fps), and `A.S.' is the overall speed (fps). The DMM-Net does not need explicit detector. DMM-Net+ uses a subset of Omni-MOT for data augmentation.}
    %\vspace{-5mm}
      \label{tab:result-ua}
      \resizebox{\textwidth}{!}{
        \begin{tabular}{lcccccccccccc}\hline
      \textbf{Tracker} & \textbf{Detector} & \textbf{PR-MOTA}$\uparrow$   & \textbf{PR-MT}$\uparrow$   & \textbf{PR-ML}$\downarrow$   & \textbf{PR-IDS}$\downarrow$ & \textbf{PR-FRAG}$\downarrow$ & \textbf{PR-FP}$\downarrow$  & \textbf{PR-FN}$\downarrow$    & \textbf{T.S.}$\uparrow$ & \textbf{D.S.}$\uparrow$ & \textbf{A.S.}$\uparrow$ \\
      \hline
      CEM~\cite{Andriyenko2011}   & ACF~\cite{Dollar2014}                 & 4.50\%             & 2.90\%            & 37.10\%           & 265.4         & 366.0         & 15180.3           & 270643.2          & 3.74      & 0.67  & 0.57                  \\
      CMOT~\cite{Zhua}            & ACF~\cite{Dollar2014}                 & 7.80\%             & 14.30\%           & 20.70\%           & 418.3         & 2161.7        & 81401.4           & 183400.2          & 3.12      & 0.67  & 0.55                  \\
      DCT~\cite{Andriyenko2012}   & ACF~\cite{Dollar2014}                 & 7.90\%             & 4.80\%            & 34.40\%           & 108.1         & 101.4         & 13059.7           & 251166.4          & 1.29      & 0.67  & 0.44                  \\
      GOG~\cite{Pirsiavash2011a}  & ACF~\cite{Dollar2014}                 & 10.80\%            & 12.20\%           & 22.30\%           & 3950.8        & 3987.3        & 45201.5           & 197094.2          & 319.29    & 0.67  & 0.67                  \\
      H2T~\cite{Wen2014}          & ACF~\cite{Dollar2014}                 & 8.20\%             & 13.10\%           & 21.30\%           & 1122.8        & 1445.8        & 71567.4           & 189649.1          & 1.08      & 0.67  & 0.41                  \\
      IHTLS~\cite{Dicle2013}      & ACF~\cite{Dollar2014}                 & 6.60\%             & 11.50\%           & 22.40\%           & 1243.1        & 4723.0        & 72757.5           & 198673.5          & 5.09      & 0.67  & 0.59                  \\
      CEM~\cite{Andriyenko2011}   & DPM~\cite{felzenszwalb2010object}     & 3.30\%             & 1.30\%            & 37.80\%           & 265           & 317.1         & 13888.7           & 270718.5          & 4.49      & 0.17  & 0.16                  \\
      DCT~\cite{Andriyenko2012}   & DPM~\cite{felzenszwalb2010object}     & 2.70\%             & 0.50\%   & 42.70\%           & 72.2          & \textbf{68.8} & 7785.8            & 280762.2          & 2.85      & 0.17  & 0.16                  \\
      GOG~\cite{Pirsiavash2011a}  & DPM~\cite{felzenszwalb2010object}     & 5.50\%             & 4.10\%            & 27.70\%           & 1873.9        & 1988.5        & 38957.6           & 230126.6          & 476.52    & 0.17  & 0.17                  \\
      IOUT~\cite{Bochinski2017d}  & DPM~\cite{felzenszwalb2010object}     & 1.92\%             & 0.84\%            & 43.70\%           & \textbf{61.4} & 106.0         & \textbf{3111.5}   & 290412.2          & 100842    & 0.17  & 0.17                  \\
      CEM~\cite{Andriyenko2011}   & R-CNN~\cite{Girshick2014}             & 2.70\%             & 2.30\%            & 34.10\%           & 778.9         & 1080.4        & 34768.9           & 269043.8          & 5.4       & 0.1   & 0.10                  \\
      DCT~\cite{Andriyenko2012}   & R-CNN~\cite{Girshick2014}             & 11.70\%            & 10.10\%           & 22.80\%           & 758.7         & 742.9         & 336561.2          & 210855.6          & 0.71      & 0.1   & 0.09                  \\
      CMOT~\cite{Zhua}            & R-CNN~\cite{Girshick2014}             & 11.00\%            & {\bf 15.70\%}           & 19.00\%           & 506.2         & 22551.1       & 74253.6           & \textbf{177532.6} & 3.59      & 0.1   & 0.10                  \\
      GOG~\cite{Pirsiavash2011a}  & R-CNN~\cite{Girshick2014}             & 10.00\%            & 13.50\%           & 20.10\%           & 7834.5        & 7401.0        & 58378.5           & 192302.7          & 352.8     & 0.1   & 0.10                  \\
      H2T~\cite{Wen2014}          & R-CNN~\cite{Girshick2014}             & 11.10\%            & 14.60\%           & 19.80\%           & 1481.9        & 1820.8        & 66137.2           & 184358.2          & 2.78      & 0.1   & 0.10                  \\
      IHTLS~\cite{Dicle2013}      & R-CNN~\cite{Girshick2014}             & 8.30\%             & 12.00\%           & 21.40\%           & 1536.4        & 5954.9        & 68662.6           & 199268.8          & 11.96     & 0.1   & 0.10                  \\ \hline
      % SSDT (M.C.) & -         & 7.45\%            & 32.43\%           & 6.75\%            & 18.78\%           & 592.8         & 690.7         & 48035.1           & 187038.2          & -         & -     & 116.64                \\
      \textbf{DMM-Net}               & -                                     & \textbf{11.80\%}   & 10.30\%           & \textbf{15.20\%}  & 230.3         & 658.0         & 36238.8           & 194886.4          & -         & -     & \textbf{123.25}       \\ 
      \textbf{DMM-Net+}              & -                                     & \textbf{12.20\%}   & 10.80\%           & \textbf{14.90\%}  & 228.2         & 674.1         & 36355.8           & 192289.6          & -         & -     & \textbf{123.25}       \\\hline   
      \end{tabular} }
   \end{center}
 \tabcolsep=1pt
 %\vspace{-5mm}
 \end{table*}
 
  %The UA-DETRAC challenge is a challenge for vehicle detection and tracking. 
 %It based on a large scale dataset, which comprises 100 videos recorded in 24 different locations.
% These scenes include urban highway, traffic crossings, and T-junctions, etc. 
 %The detail of the UA-DETRAC dataset is summarized in Tab.~\ref{tab:compare}.
%  The UA-DETRAC challenge recently released the testing dataset and evaluation utils. 
%The result of submitted approaches is evaluated on the remote server.
 
 % dataset
 %The videos in UA-DETRAC dataset has a consistent frame size of $540 \times 960$, and a frame rate of 25 fps.
 %All the videos are recorded with static cameras. The large size of data, the variety of the scenes, and the junctions make the tracking in this dataset a challenging task. 

 % metrics
 %The evaluation metrics used by UA-DETRAC are based on CLEAR MOT and MT/ML. The prefix ``PR' indicates that the Precision-Recall curve of the detection computes these metrics.
 %For example, to compute the PR-MOTA, the thresholds of the detectors are gradually varied to calculate a 2D precision-recall curve. 
 %For each point on the PR curve, the MOTA is computed to get a 3D curve. The PR-MOTA is computed as the integral score of the resulting curve.
 %It is the same to compute other metrics in UA-DETRAC metrics.
 
 % result
 
 % \begin{figure}[t]
 %     \centering
 %     \includegraphics[width=3in]{imgs/ua_loss.png}
 %     \caption{The losses of training FMMN on UA-DETRAC dataset for 35 epoches.}
 %     \label{fig:ua_loss}
 % \end{figure}
 
 Notice that our technique does not require an external `detector' and achieves highly promising PR-MOTA score for UA-DETRAC @ 120+ fps - orders of magnitude faster than the existing methods.
 In the \SJ{third} row of Fig.~\ref{fig:ua_omot_examples}, we show an example of tracking result. The box color indicates the predicted object identity.  Our tracker leverages information from previous frames to detect partially occluded objects, e.g.~vehicle-3 in frame 518. The figure also shows the generalization power of our in-built detection mechanism. For instance, vehicle-1 is a very rare vehicle that is consistently assigned the correct identity. In the supplementary material, we provide further example tracking videos that clearly demonstrate successful tracking by DMM-Net for UA-DETRAC.

 \section{Conclusion}\label{sec:c}
 %\vspace{-3mm}
 In the context of tracking-by-detection, we proposed a deep network DMM-Net that removes the need for an explicit external detector and performs tracking at 120+ fps to achieve 12.80\% PR-MOTA value on the UA-DETRAC challenge. Our network meticulously models object motions, \SJ{classes} and their visibility that are subsequently used for efficiently associating the object tracks. The proposed network provides an end-to-end solution for detection and tracklet generation across multiple video frames. 
 We also propose a realistic CARLA simulator based large-scale dataset with over 14M frames for vehicle tracking. The dataset provides precise and comprehensive ground truth with full control over data parameters, which  allows for the much needed transparency in  evaluation. We also provide scripts to generate more data under our framework and we make the code of DMM-Net public.

\clearpage
% ---- Bibliography ----
%
% BibTeX users should specify bibliography style 'splncs04'.
% References will then be sorted and formatted in the correct style.
%
\bibliographystyle{splncs04}
\bibliography{ms}
\end{document}

% \renewcommand\thelinenumber{\color[rgb]{0.2,0.5,0.8}\normalfont\sffamily\scriptsize\arabic{linenumber}\color[rgb]{0,0,0}}
% \renewcommand\makeLineNumber {\hss\thelinenumber\ \hspace{6mm} \rlap{\hskip\textwidth\ \hspace{6.5mm}\thelinenumber}}
% \linenumbers
\pagestyle{headings}
\mainmatter
\def\ECCVSubNumber{4729}  % Insert your submission number here

\title{Simultaneous Detection and Tracking with Motion Modelling for Multiple Object Tracking  \\ (Supplementary Material)} % Replace with your title

% INITIAL SUBMISSION 
\begin{comment}
\titlerunning{ECCV-20 submission ID \ECCVSubNumber} 
\authorrunning{ECCV-20 submission ID \ECCVSubNumber} 
\author{Anonymous ECCV submission}
\institute{Paper ID \ECCVSubNumber}
\end{comment}
%******************

% CAMERA READY SUBMISSION
% \begin{comment}
\titlerunning{Deep Motion Modeling Network for MOT}
% If the paper title is too long for the running head, you can set
% an abbreviated paper title here
%
\author{ShiJie Sun\inst{1} \and
Naveed Akhtar\inst{2} \and
XiangYu Song\inst{3} \and
HuanSheng Song\inst{1} \and
Ajmal Mian\inst{2} \and
Mubarak Shah\inst{4}
}
\authorrunning{S. Sun, N. Akhtar, et al.}
% First names are abbreviated in the running head.
% If there are more than two authors, 'et al.' is used.
%
\institute{
Chang'an University, Xi'an, Shaanxi, China \\
\email{\{shijieSun,hshsong\}@chd.edu.cn} \and
University of Western Australia, 35 Stirling Highway, Crawley, WA, Australia \\
\email{\{naveed.akhtar,ajmal.mian\}@uwa.edu.au} \and
%\url{http://www.springer.com/gp/computer-science/lncs} \and
Deakin University, RWaurn Ponds, Victoria 3216, Melbourne, Australia \\
\email{xiangyu.song@deakin.edu.au}\and
University of Central Florida, Orlando, FL, America \\
\email{shah@crcv.ucf.edu}
}
% \end{comment}
%******************
\maketitle

\appendix

\section{Public Source Code \& Dataset}
Along with the submitted manuscript, we provide the source code of the proposed DMM-Net and publish the proposed Omni-MOT dataset. We also provide our implementation to generate more videos similar to the proposed dataset using the CARLA simulator~\cite{Koltun2017}. Below, the links are provided for anonymous repositories for the sake of the review process. The links will be made public after the acceptance. (Click on the highlighted text to open the URL). 
\begin{itemize}
   \item \href{https://github.com/shijieS/DMMN}{DMM-Net} is the source code of DMM-Net. It also contains the training and testing script for both UA-DETRAC~\cite{Wen2015a} and Omni-MOT dataset, and instructions for reproducing the result of our methods.
   \item \href{https://github.com/shijieS/OmniMOTDataset}{Omni-MOT Dataset} provides the link to the dataset along with the related description.
   \item \href{https://github.com/shijieS/OMOTDRecorder}{Omni-MOT Script} is the source code for generating the Omni-MOT dataset and extending it. It includes the script for recording the MOT data and playing the recorded dataset. 
\end{itemize}

\section{Videos for Dataset and Further Results}
We provide the following videos for the review process. These and further videos will also be made public after the acceptance:
\begin{itemize}
    \item \href{https://www.dropbox.com/sh/nom2u2s7snivwhu/AABCywWoePDi0MnNXM9tXzyfa?dl=0&preview=omni-mot_dataset.mp4}{Omni-MOT dataset videos} illustrates different scenes,  camera viewpoints and weather conditions used in the generated large-scale realistic dataset.
    \item \href{https://www.dropbox.com/sh/nom2u2s7snivwhu/AABCywWoePDi0MnNXM9tXzyfa?dl=0&preview=omni_result.avi}{Further results on Omni-MOT} illustrates more tracking results on the proposed dataset.
    \item \href{https://www.dropbox.com/sh/nom2u2s7snivwhu/AABCywWoePDi0MnNXM9tXzyfa?dl=0&preview=UA-DETREAC-MVI_39271.avi}{Further results on UA-DETRAC} show tracking results on a representative scene from the UA-DETRAC challenge.
\end{itemize}

\section{Further  Details of the Dataset}
\begin{figure}[t]
   \centering
   \includegraphics[width=4in]{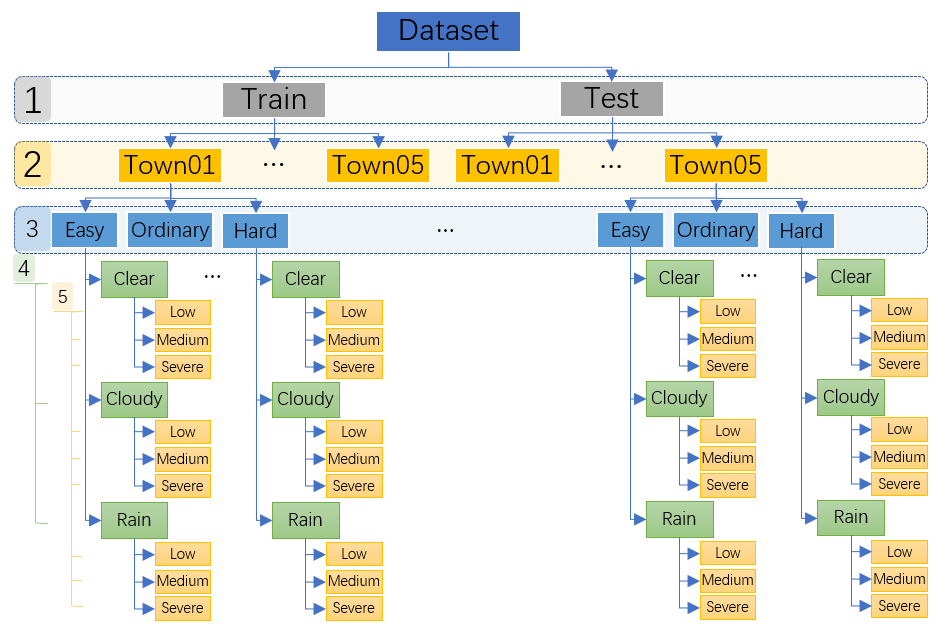}
   \caption{Overall structure of the Omni-MOT dataset.}
   \label{fig:different_levels}
\end{figure}

The structure of the proposed dataset can be best understood under five dimensions of divisions. We depict these dimensions as different levels of a block diagram in Fig.~\ref{fig:different_levels} for a clear overview. Along the first dimension, we split the dataset into the training and testing sets. Along the second, five towns of the CARLA simulator are employed for making the dataset diverse. For each town, we set the camera with different viewpoints for the third dimension of division. These viewpoints include three levels of difficulty, namely, Easy, Ordinary, and Hard level. Along the fourth dimension, different weather conditions split the data. These weather conditions contain Clear, Cloudy, and Rainy weather. The last variability that makes our data diverse consists of three congestion levels, namely; Low, Medium, and Severe congestion. 
Details are given below. 

 \vspace{1mm}
 \noindent{\bf Train/Test split}: {The training set consists of $1,755$ videos, $8,775$K frames, $134.2$K tracks, and $68.88$M boxes. The testing set includes $1,755$ videos, $5,265$K frames, $122.37$K tracks and $41.36$M boxes.}
 
 \vspace{1mm}
 \noindent{\bf Towns}: {There are five towns in our dataset, whose details are given in Table~\ref{tab:CityInformation}. Among these, Town05 is the largest city that also has three overpass roads. Town02 is the smallest city, whereas Town03 also contains a tube. Town04 is the most populous in terms of T-junctions.}
 
 \vspace{1mm}
 \noindent{\bf Camera viewpoints}: {39 cameras are placed in each city with different viewpoints. The camera horizontal field of view is $90^{\circ}$. We refer to the \href{https://www.dropbox.com/sh/nom2u2s7snivwhu/AABCywWoePDi0MnNXM9tXzyfa?dl=0&preview=omni-mot_dataset.mp4}{Omni-MOT dataset videos} to visually observe the viewpoints.}
 
  \vspace{1mm}
 \noindent{\bf Weather conditions}: {Three kinds of weather are simulated, namely, Clear, Cloudy, and Rainy, by changing the weather parameters of the CARLA simulator. These weather parameters include cloudiness and precipitation, and their values range from 0 to 100. The cloudiness of  Clear is 15, and the cloudiness of Cloudy and Rainy are 80. Both the precipitation of Clear and Cloudy are 0, and the precipitation of Rainy is 60.}

 \vspace{1mm}
 \noindent{\bf Road congestion}:  We include three levels of traffic congestion, i.e.~low, medium, and severe congestion. Because cities have different sizes, these congestion levels are decided by different numbers of vehicles. Table~\ref{tab:congestion} summarizes the number of vehicles for the chosen level of congestion in all five towns.

In the proposed dataset, five different simulated cities are considered. For each city, we use up to 39 cameras. The cameras are placed with viewpoints that have three levels of difficulty for the MOT scenarios. Namely, (a) Easy view: which results in no occlusion of the vehicles. (b) Ordinary view: that allows temporary occlusions but forbids continuous occlusions. (c) Hard view: that allows continuous occlusions in the videos. Collectively, we provide $90$ scenes in the dataset that result in $3,510$ videos. There are $14.04$M frames of size $1920 \times 1080$ in the OMOT dataset that are recorded in the `XDIV' format to provide high-quality videos with acceptable memory size.

%The detail of these five cities are given in Tab.~\ref{tab:CityInformation}. In these cities, Town05 is the largest city and has three overpass roads, the size of Town01 and Town02 are smaller than those of other cities, and Town03 and Town04 contain the most T-junction.

\begin{table}[t]
   \begin{center}
       \small
       \caption{Details of five towns. Column ``Size'' is manually measured and its format is $\mathop{Width} \times \mathop{Length}$}
       \label{tab:CityInformation}
       \tabcolsep=3pt
       \begin{tabular}{ccccccc}\hline
       Name & Size (m)                     & Cross & T-junction & Roundabout & Overpass & Tunnel\\
       \hline
       Town01    & $342\times 413$         & 0     & 12         & 0          & 0        & 0      \\
       Town02    & $205\times 208$         & 0     & 9          & 0          & 0        & 0      \\
       Town03    & $438\times 483$         & 5     & 14         & 2          & 0        & 1      \\
       Town04    & $816\times 914$         & 8     & 21         & 3          & 1        & 0      \\
       Town05    & $430\times 486$         & 13    & 8          & 0          & 3        & 0      \\\hline
       \end{tabular}
   \end{center}
\end{table}

\begin{table}[t]
   \small
   \centering
   \caption{The number of vehicles for each congestion level of different towns.}
   \label{tab:congestion}
   \begin{center}
      \begin{tabular}{cccc}
      \hline
      City & Low & Medium & Severe \\ \hline
      Town01    & 50     & 75     & 95     \\
      Town02    & 50     & 75     & 95     \\
      Town03    & 100    & 170    & 230    \\
      Town04    & 100    & 170    & 230    \\
      Town05    & 100    & 170    & 230    \\ \hline
      \end{tabular} 
   \end{center}
   \end{table}

\subsection{Comparison to the existing MOT datasets}
To put the proposed dataset into a better perspective, we also compare it to the existing popular datasets for the MOT task. In Table~\ref{tab:compare}, we provide the comparison. Our dataset comprises 3,510 videos, 14M+ frames, 250K tracks, and 110M+ bounding boxes, whose frame number is almost 1,200 times larger than MOT17. 
The number of provided tracks and boxes are 210 and 30 times larger than UA-DETRAC. Besides, for the proposed Omni-MOT, all the boxes and tracks are automatically generated by the enumerator that avoids any labeling error.
In the table, we include nuScenes~\cite{caesar2019nuscenes} and Waymo~\cite{sun2019scalability} for the sake of comprehensive benchmarking. Nevertheless, these datasets are related to self-driving vehicles that are captured with moving cameras. 

\begin{table}[t]
   \begin{center}
      \small
      \caption{Comparison with other popular MOT datasets. Columns ``Frames`` is the  number of frames 
   ($1k=10^3, 1M=10^6$), ``Tracks`` is the number of tracks and ``Boxes`` is the number of bounding boxes. "-" indicates that no information is provided.}
      \label{tab:compare}
      \tabcolsep=3pt
      \begin{tabular}{lcccccc}
      \hline
      \multirow{2}{*}{Dataset} & \multicolumn{3}{c|}{Training} & \multicolumn{3}{c}{Testing} \\
      \cline{2-7}
                               & Frames  & Tracks   & Boxes   & Frames & Tracks   & Boxes   \\\hline
      PETS~\cite{Ferryman2009}                     & -       & -        & -       & 1.5k   & 106      & 18.5k   \\
      KITTI~\cite{Geiger2012}                    & 8k      & -        & -       & 11k    & -        & -       \\
      TUD~\cite{andriluka2008people}                      & 610     & -        & 610     & 451    & 31       & 2.6k    \\
      MOT15~\cite{Leal-Taixe2015}                    & 5.5k    & 500      & 39.9k   & 5.8k   & 721      & 61k     \\
      MOT17~\cite{MilanL0RS16}                    & 5.3k    & 467      & 110k    & 5.9k   & 742      & 182k    \\
      UA-DETRAC~\cite{Wen2015a}                & 84k     & 5.9k     & 578k    & 56k    & 2.3k     & 632k    \\
      nuScenes~\cite{caesar2019nuscenes}  & 40k & - & 1.4M & - & - & - \\
      Waymo~\cite{sun2019scalability}                 & 154k    & -         &  8.6M       & 23k    &     -     & 1.3M \\
      Omni-MOT(Ours)              & \textbf{8775k}    & \textbf{134.2k}   & \textbf{68.88M}   & \textbf{5265k}  & \textbf{122.37k}         &  \textbf{41.36M}       \\ \hline
      \end{tabular}
   \end{center}
\end{table}

\subsection{On the ground-truth annotations}
The ground-truth annotations is generated by the CARLA simulator, which allows us to capture comprehensive information on the target objects with high precision. Hence, besides being accurate, our ground truth annotations are detailed enough to be used for other related problems, such as 3D estimation, velocity estimation, camera calibration, etc. To this end, we bring as much information as we can into the ground truth file. 
Table~\ref{tab:gt} gives details of the format of the ground truth files ({available through the dataset download link provided above}).
The ``3D bbox'' at columns 6-13 contains values describing point coordinates. 
These points are the image projection of a minimum 3D cuboid envelope of the vehicle in the world coordinates. 
The column ``bbox'' is calculated by the minimum rectangle envelope of these points.
On index 17, ``integrity'' encodes the visibility of vehicles. A clear description of the remaining entities is provided in the table. 

%%%%%%%%%%%%%%%% Move from the main paper%%%%%%%%%%%%%%%%%%%% 
% Besides controlling the camera viewpoints, we include three different weather conditions; clear, cloudy and rainy. We also include three levels of traffic congestion in the OMOT dataset i.e. low, medium and severe congestion. We provide further details of the dataset in the supplementary material, where we  also include a video showing different scenes, camera viewpoints and weather conditions. CARLA simulator allows us to capture comprehensive information on the target objects with high precision. Hence, besides being accurate, our ground truth annotations are detailed enough to be used for other related problems, e.g.~velocity estimation, 3D localization, camera calibration. 
% The ground-truth annotations is generated by the CARLA simulator, which allows us to capture comprehensive information on the target objects with high precision. Hence, besides being accurate, our ground truth annotations are detailed enough to be used for other related problems, e.g.~velocity estimation, 3D localization, camera calibration. 
%%%%%%%%%%%%%%%%%%%%%%%%%%%%%%%%%%%%%%%%%%%%%%%%%%%%%%%%%%%% 
 
%Moreover, we use the video (``\textit{omni-mot\_dataset.mp4}'') to briefly show the scenes and ground truth of Omni-MOT dataset.
\begin{table}[t]
   \centering
   \begin{center}
   \small
   \caption{Data format of the ground truth file provided with the dataset.}
   \label{tab:gt}
   \resizebox{\textwidth}{!}{
   \begin{tabular}{cp{2.7cm}p{9.5cm}}
   \hline
   Index & Name                & Description                                                             \\ \hline
   0        & frame index         & 0-based frame index                                                     \\
   1        & vehicle id          & Unique ID of vehicle (0-based)                                      \\
   2        & bbox                & Represents left, top, right, and bottom of the vehicle bounding box       \\
   6        & 3d bbox             & The 8 points of the vehicle’s 3D bounding boxes in image coordinates \\
   14       & vehicle position    & Vehicle’s center in the world coordinates    \\
   17       & integrity           & Integrity of the vehicle. Binary value  in (0, 1)                        \\
   18       & velocity vector     & Velocity vector in the world coordinates                            \\
   21       & acceleration vector & Acceleration vector in the world coordinates                        \\
   24       & wheel number        & Number of vehicle wheels                                                        \\
   25       & camera view size    & The width and the height of the camera view                                 \\
   27       & camera FOV          & The field of view of the camera                                         \\
   28       & camera position     & Camera position in the world coordinates                           \\
   31       & camera rotation     & Camera rotation in the world coordinates                                              \\
   34       & weather condition   & Weather condition in the current frame.                                 \\ \hline
   \end{tabular}}
   \end{center}
\end{table}

%  \newpage
\section{Anchor Tubes}
\begin{figure}[t]
   \centering
   \includegraphics[width=3in]{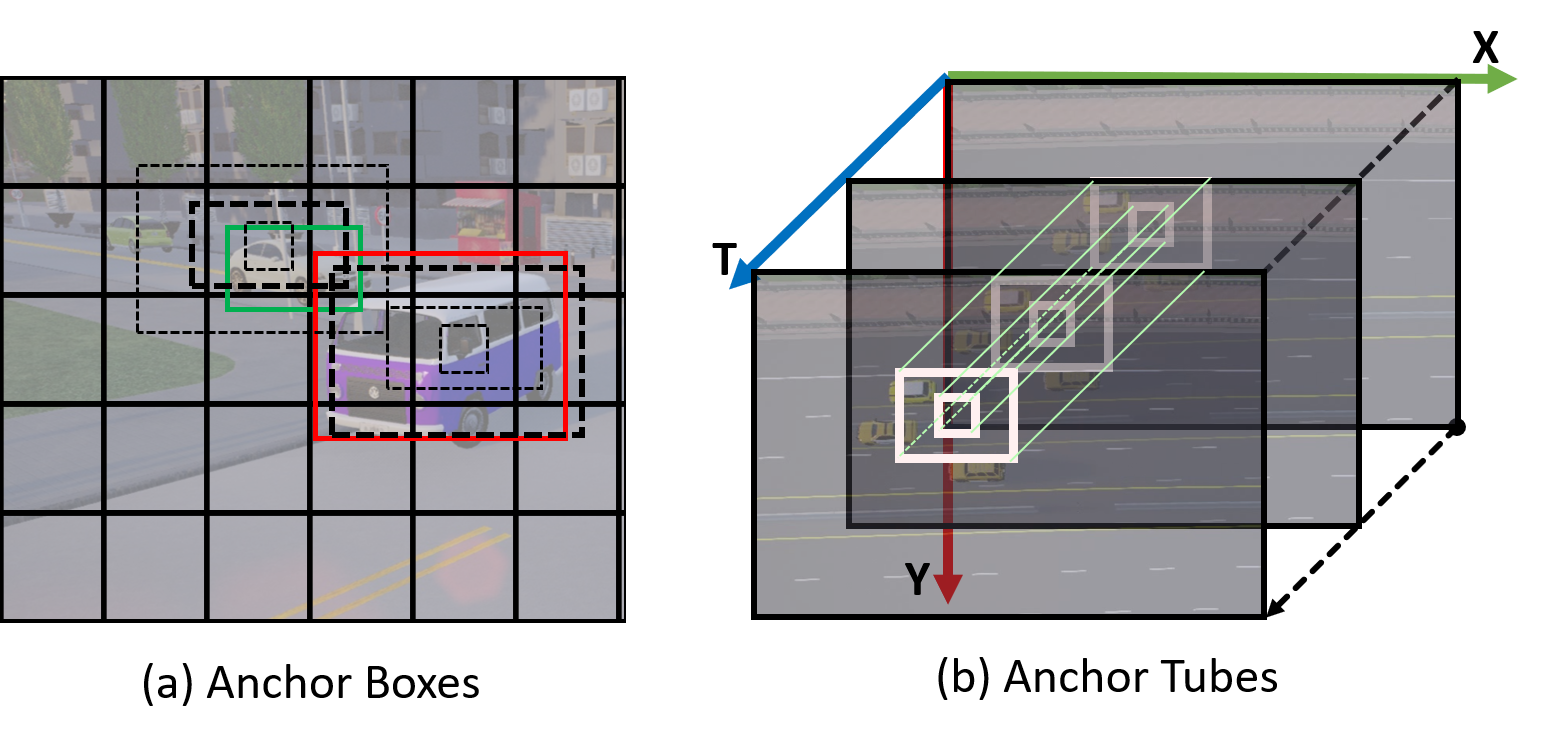}
   \caption{Illustration of the anchor tubes. The anchor boxes in (a) are a set of boxes in the 2D coordinates. The SSD model predicts the scaling and translation parameters relative to each anchor box. Extended from anchor boxes, the anchor tubes in (b) are a set of cuboids in the 3D coordinate. Each anchor tube consists of $N_F$ boxes. Our proposed network  predicts the tube shape offset parameters along the temporal dimension, the confidence for each object class, and the visibility of each box in the tube.}
   \label{fig:anchor_tunnels}
\end{figure}

The proposed notion of anchor tube is an extension of the concept of anchor boxes in SSD~\cite{Liu2016}. 
%The anchor tube is extended from anchor boxes of SSD.
In our technique, the anchor tubes are pre-defined and distributed at every position of the selected feature maps. An anchor tube is essentially a set of anchor boxes that share the same location in multiple frames along the temporal dimension, as illustrated in Fig.~\ref{fig:anchor_tunnels}. The  Fig.~\ref{fig:anchor_tunnels}(a)  depicts three pre-defined anchor boxes at each position of a 3-D feature map. The Fig.~\ref{fig:anchor_tunnels}(b) illustrates a  pre-defined anchor tube at a position of a 4-D feature map. 
Similar to the main goal of the anchor boxes, the anchor tubes format the network output dimensions. Consequently, our network is designed to predict the tube shape offset parameters along the temporal dimension, the confidence for each object class, and the visibility of each box in the tube.

% Because of different input frames containing a various number of ground boxes, the SSD network cannot output these ground truth boxes directly. They encode the ground truth box into the anchor box to make the dimension of ground truth consistent. Similar to SSD, we use the same techniques. The number of ground truth tracks varies at different input video frames. Therefore, we need to encode these tracks into the anchor tunnels to make the output's dimension of the network constant. 

% Fig.~\ref{fig:anchor_tunnels} also illustrates the difference between anchor boxes and anchor tunnels. Similar to anchor boxes, the anchor tunnels target to formatting the output of the network. 
%  For each anchor tunnel, our network predicts the shape offset parameters relative to time, the confidence for each category, and the visibility of each box in the anchor tunnel. 

 \section{Further Details on Motion Model}
 The proposed DMM-Net outputs encoded anchor tubes directly. However, it entails predicting numerous parameters (unless a compact encoding for the tubes is used). For instance, assume that an anchor tube contains 16 boxes. In this case, the network would need to output 16x4 scalar values to describe the tube. To limit the output parameters, we introduce the motion function to describe these boxes in an encoded anchor tube. In our experiments, we use the quadratic function that only needs 3x4 motion parameters to describe an encoded anchor tube. 
 The Eq.~\eqref{eq:final_model} below states the relationship between the motion parameters and the ground truth tracks.
 % , we introduce the . It describes the relationship between the box $(b_{i,t}^{cx}, b_{i,t}^{cy}, b_{i,t}^{w}, b_{i,t}^{h})$ of the $i^{th}$ best-matched track and the estimated motion parameters at the $t^{th}$ frame.
 %---------------------

 \begin{equation}
   \label{eq:final_model}
   \left\{\begin{matrix}
   b_{i,t}^{w}  = a_{i,t}^{w}\mathop{exp}({p_{11} t^2 + p_{12}t + p_{13} + \Delta^{w}}) \\ 
   b_{i,t}^{h}  = a_{i,t}^{h}\mathop{exp}({p_{21} t^2 + p_{22}t + p_{23} + \Delta^{h}}) \\ 
   b_{i,t}^{cx} = p_{31}a_{i,t}^{w} t^2 + p_{32}a_{i,t}^{w} t + p_{33}a_{i,t}^{w} + a_{i,t}^{cx} + \Delta^{cx} \\ 
   b_{i,t}^{cy} = p_{41}a_{i,t}^{h} t^2 + p_{42}a_{i,t}^{h} t + p_{43}a_{i,t}^{h} + a_{i,t}^{cy} + \Delta^{cy}, 
   \end{matrix}\right.
\end{equation}
where $(a_{i, t}^{cx}, a_{i, t}^{cy}, a_{i, t}^{w}, a_{i, t}^{h})$ is the pre-defined box (center x, center y, width, height) of $i^{th}$ anchor tube at $t^{th}$ frame, $(b_{i,t}^{cx}, b_{i,t}^{cy}, b_{i,t}^{w}, b_{i,t}^{h})$ is the box of $i^{th}$ ground truth track at $t^{th}$ frame, $\{p_{11}, \cdots, p_{43}\}$ represents the motion parameters of the $i^{th}$ encoded anchor tube, and $(\Delta^{cx}, \Delta^{cy}, \Delta^{w}, \Delta^{h})$ is the localization error of our network output. 
From Eq.~\eqref{eq:final_model}, we can see that the box center $(b_{i,t}^{cx}, b_{i,t}^{cy})$ of the ground truth track is a quadratic function of time, while $(b_{i,t}^{w}, b_{i,t}^{h})$ is more a complicated function. The motion parameters are able to successfully model object motion for short time slots to generate effective tracklets.

%  \begin{figure}[t]
%    \centering
%    \includegraphics[width=5in]{imgs/motion_model.png}
%    \caption{...}
%    \label{fig:motion_model}
% \end{figure}

\begin{table*}[t]
   \begin{center}
      \small
         \caption{Further results on Omni-MOT with Medium and Hard camera views and Cloudy weather conditions: The symbol $\uparrow$ indicates higher values are better, and $\downarrow$ implies lower values are favored. }
         \label{tab:result-omot}
      \tabcolsep=1pt
   \resizebox{\textwidth}{!}{
   \begin{tabular}{cccccccccccccccccc}
   \hline
   \textbf{Type} & View & \textbf{Camera} & \textbf{IDF1}$\uparrow$ & \textbf{IDP}$\uparrow$ & \textbf{IDR}$\uparrow$ & \textbf{Rcll}$\uparrow$ & \textbf{Prcn}$\uparrow$ & \textbf{GT}$\uparrow$ & \textbf{MT}$\uparrow$ & \textbf{PT}$\uparrow$ & \textbf{ML}$\downarrow$ & \textbf{FP}$\downarrow$ & \textbf{FN}$\downarrow$ & \textbf{IDs}$\downarrow$ & \textbf{FM}$\downarrow$ & \textbf{MOTA}$\uparrow$ & \textbf{MOTP}$\uparrow$ \\ \hline
   \multirow{3}{*}{Test} & Hard & Camera\_1 & 35.8\% & 41.8\% & 31.3\% & 55.8\% & 74.6\% & 116 & 9 & 34 & 73 & 3359 & 7801 & 78 & 184 & 36.3\% & 73.2\% \\
    & Hard & Camera\_12 & 32.0\% & 38.6\% & 27.3\% & 48.0\% & 67.7\% & 136 & 1 & 38 & 97 & 2948 & 6704 & 113 & 188 & 24.2\% & 70.5\% \\
    & Ordinary & Camera\_9 & 47.7\% & 51.1\% & 44.7\% & 68.5\% & 78.3\% & 61 & 16 & 32 & 13 & 2398 & 3986 & 42 & 150 & 49.2\% & 73.5\% \\ \hline 
   \multirow{3}{*}{Train} & Hard & Camera\_0 & 44.5\% & 48.2\% & 41.4\% & 63.7\% & 74.1\% & 137 & 17 & 36 & 84 & 2979 & 4865 & 66 & 155 & 40.9\% & 75.8\% \\
    & Hard & Camera\_1 & 30.7\% & 36.8\% & 26.3\% & 49.7\% & 69.7\% & 148 & 5 & 38 & 105 & 3236 & 7537 & 117 & 202 & 27.4\% & 73.2\% \\
    & Ordinary & Camera\_5 & 68.9\% & 70.9\% & 66.9\% & 81.5\% & 86.3\% & 94 & 32 & 40 & 22 & 3443 & 4957 & 62 & 222 & 68.3\% & 81.1\% \\ \hline
    Average & \multicolumn{2}{|c|}{-} & 47.0\% & 52.0\% & 42.8\% & 63.5\% & 77.3\% & 692 & 80 & 218 & 394 & 18363 & 35850 & 478 & 1101 & 44.4\% & 76.1\% \\ \hline
   \end{tabular}}
   \end{center}
   \vspace{-3mm}
   \end{table*}

\section{Further Quantitative Results}
To better evaluate our technique and putting the values reported in the paper into a better perspective, we provide further results of DMM-Net on two additional viewpoints Omni-MOT. These results are reported in Table~\ref{tab:result-omot}. The experiments are conducted for cloudy weather conditions. The selected scenes are from Town 05 (with 230 vehicles) that are indexed 1, 9, and 12 in the dataset, where scene-1 and scene-12 are with hard camera view, and scene-9 is with ordinary camera view. Similar to the experiments in the paper, we train the network for 22 epochs and use the same evaluation matrices as used in the paper.  The results show good performance of DMM-Net on the realistic dataset with accurate ground-truth. 
For these experiments, we observed that our network was often able to track vehicles that are fully occluded. This is a direct benefit of using motion modeling for tracking. On the flip side, we also observed a slight drift of the bounding boxes for stationary objects due to the amplification of motion caused by noisy detection. Nevertheless, this problem was never observed to cause critical problems. These observations can be verified in the videos provided on the URL links above.  

% \begin{figure}[h]
%    \centering
%    \includegraphics[width=2.5in]{imgs/amot_loss.png}
%    \caption{The loss result of OMOT dataset.}
%    \label{fig:omot_loss}
% \end{figure}
% Fig.~\ref{fig:omot_loss} demonstrates the loss of this network during the training phase.
% At the beginning of training, the loss is about 781, while the loss reduces to 5.2 in the end. 

%  \cite{Andriyenko2011}
%  \section{Acknowledgement}
%  This research was supported by 
%  ARC grant DP160101458 and DP190102443, 
%  the National Science Foundation of China (Grant No. 61572083), 
%  the Joint Found of of Ministry of Education of China (Grant No. 6141A02022610), 
%  social development projects in shaanxi province (2019SF-258),
%  High-level Perceptual Features of Near-regular Texture and Their Applications (Grant No. 61806023)
% \newpage
% \newpage
% \newpage

\clearpage
% ---- Bibliography ----
%
% BibTeX users should specify bibliography style 'splncs04'.
% References will then be sorted and formatted in the correct style.
%
\bibliographystyle{splncs04}
\bibliography{supplement}